%% file: main.tex
\documentclass[]{fairmeta}

\PassOptionsToPackage{pagebackref}{hyperref}
\input{preamble}

\usepackage{amssymb}

\title{\nymeriapp: Enriching Nymeria Dataset with Additional Annotations and Data}

\author[*]{Daniel DeTone}
\author[*]{Federica Bogo}
\author[*]{Eric-Tuan Le}
\author{Duncan Frost}
\author{Julian Straub}
\author{Yawar Siddiqui}
\author{Yuting Ye}
\author{Jakob Engel}
\author{Richard Newcombe}
\author{Lingni Ma}

\contribution[*]{Equal Contribution}
\affiliation[]{Meta Reality Labs Research}

\input{sections/0-abstract}

\date{\today}
\correspondence{Lingni Ma (\email{lingni.ma@meta.com})}

\metadata[Code]{\url{https://github.com/facebookresearch/nymeria_dataset}}
\metadata[Data]{\url{https://explorer.projectaria.com/nymeria}}

\begin{document}

\maketitle

\input{sections/1-intro}
\input{sections/2-related-work}
\input{sections/3-nymeria-plus-plus}
\input{sections/5-conclusion}

\clearpage
\newpage
\bibliographystyle{assets/plainnat}
\bibliography{main}

\clearpage
\newpage
\beginappendix

\input{sections/app1-category.tex}
\input{sections/acknowledgement.tex}

\end{document}

%% file: preamble.tex
\usepackage{xspace}

\newcommand{\nymeria}{Nymeria\xspace}
\newcommand{\nymeriapp}{NymeriaPlus\xspace}

\usepackage{tikz}

\renewcommand*{\backref}[1]{}
\renewcommand*{\backrefalt}[4]{%
    \ifcase #1%
    \else
        {\small ~(#2)}%
    \fi
}

\crefname{section}{Sec.}{Secs.}
\Crefname{section}{Sec.}{Secs.}
\crefname{figure}{Fig.}{Figs.}
\Crefname{figure}{Fig.}{Figs.}
\crefname{table}{Tab.}{Tabs.}
\Crefname{table}{Tab.}{Tabs.}
\crefname{equation}{Eq.}{Eqs.}
\Crefname{equation}{Eq.}{Eqs.}

%% file: sections/0-abstract.tex
\abstract{

  \def\iw{14.3em}
  \begin{tikzpicture}[inner sep=1pt]
    
    \node(p01){\includegraphics[width=\iw]{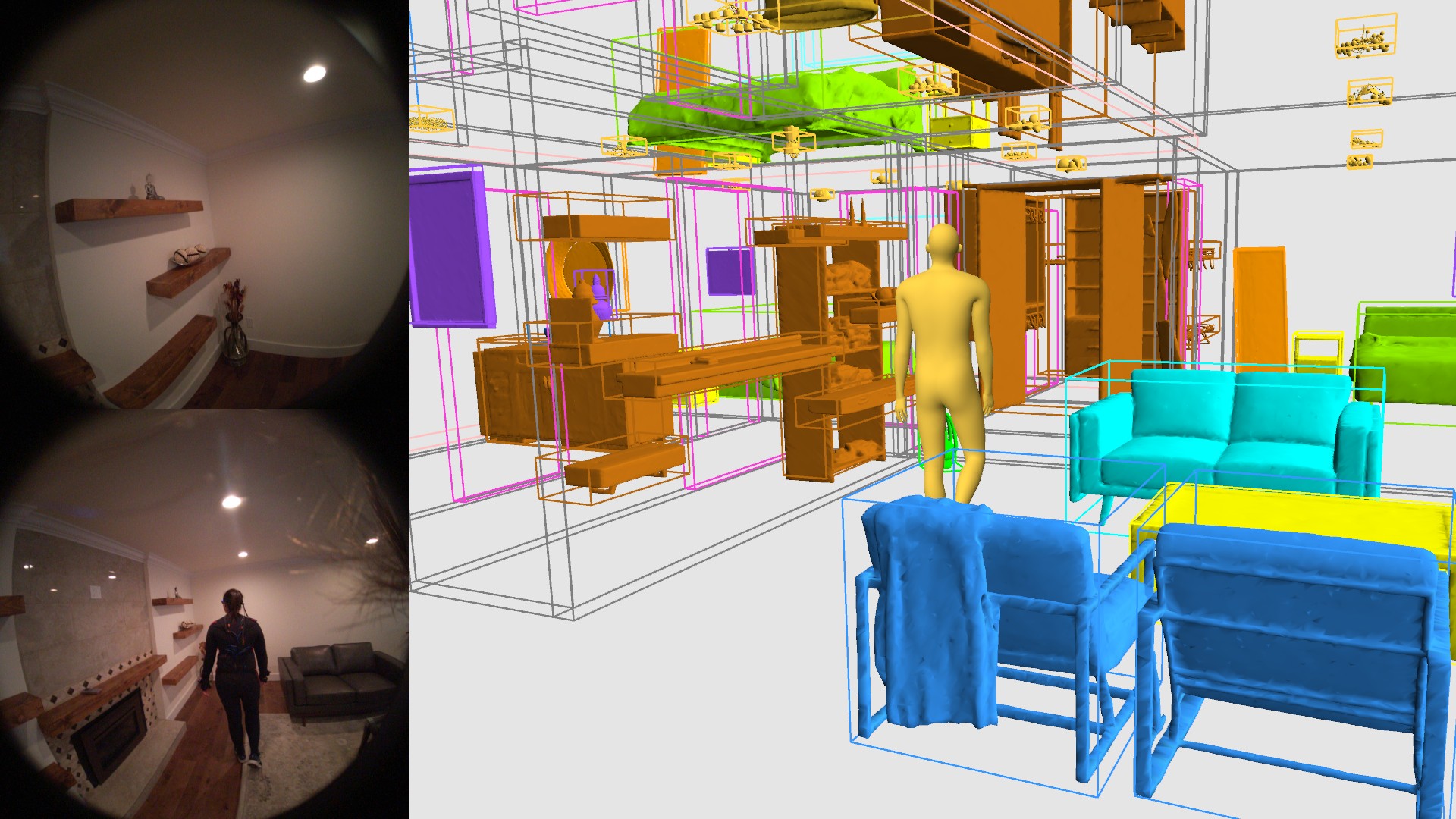}};
    \node(p02) at (p01.east) [anchor=west]{\includegraphics[width=\iw]{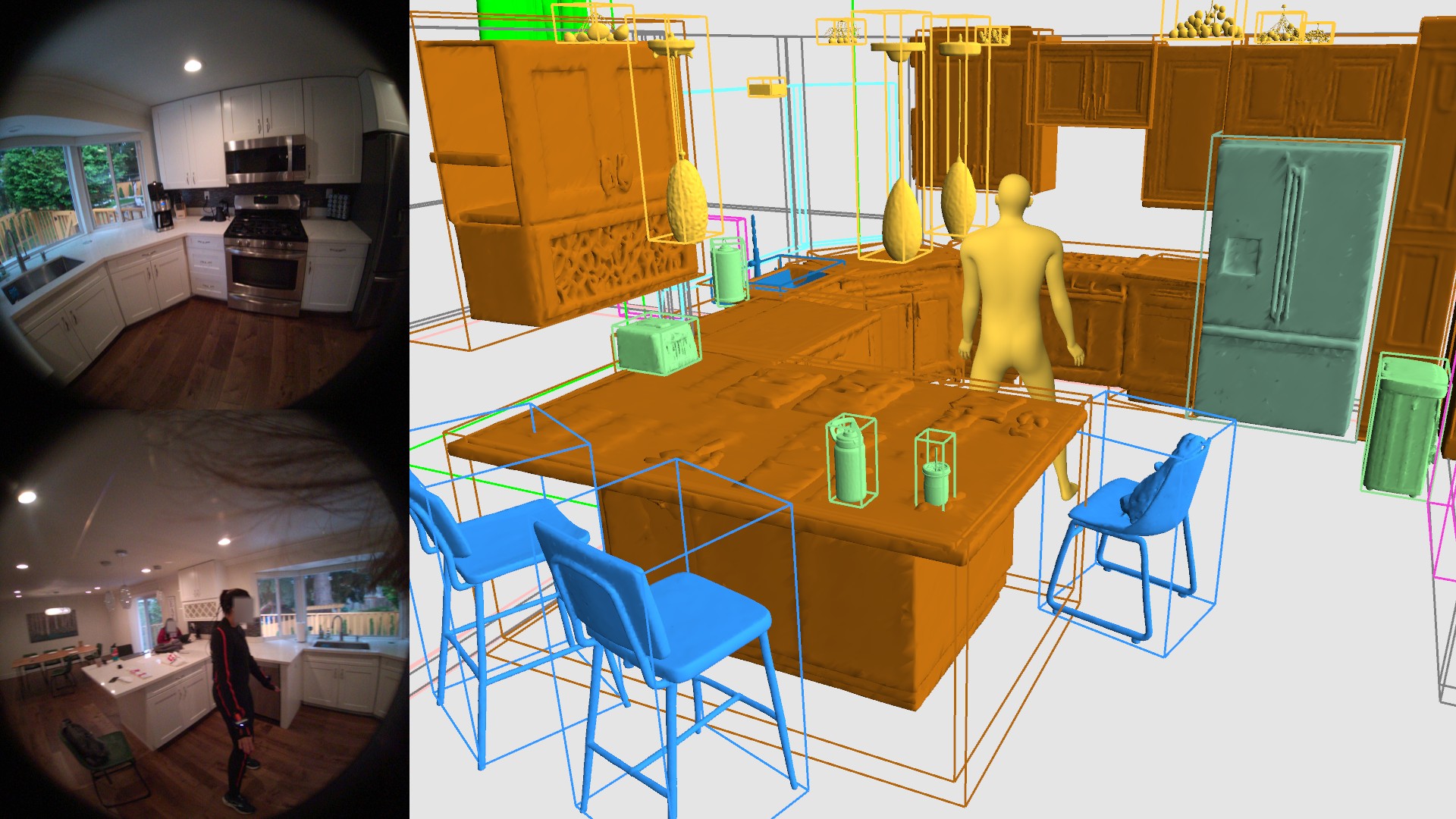}};
    \node(p03) at (p02.east) [anchor=west]{\includegraphics[width=\iw]{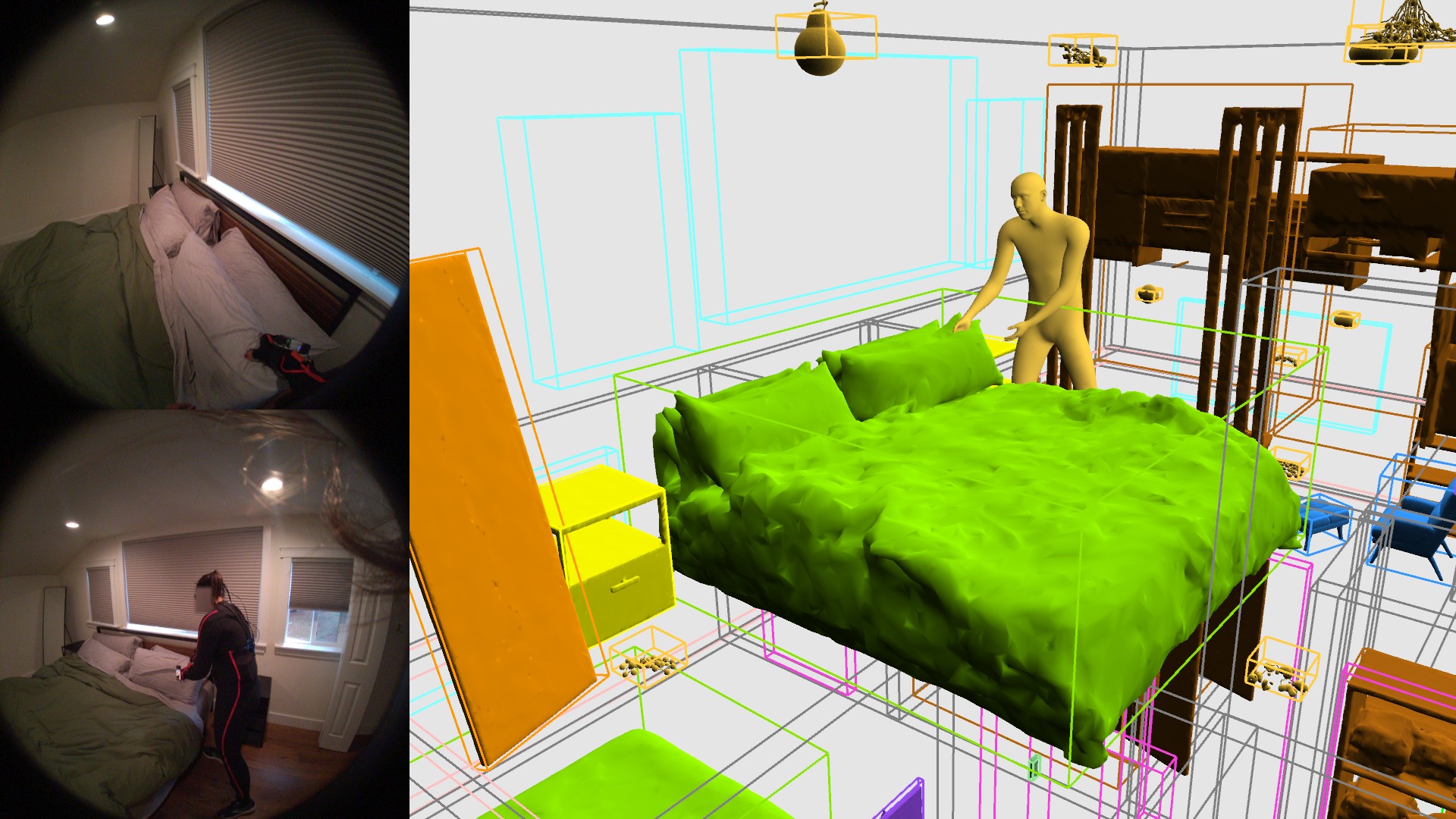}};

    \node(p11) at (p01.south) [anchor=north]{\includegraphics[width=\iw]{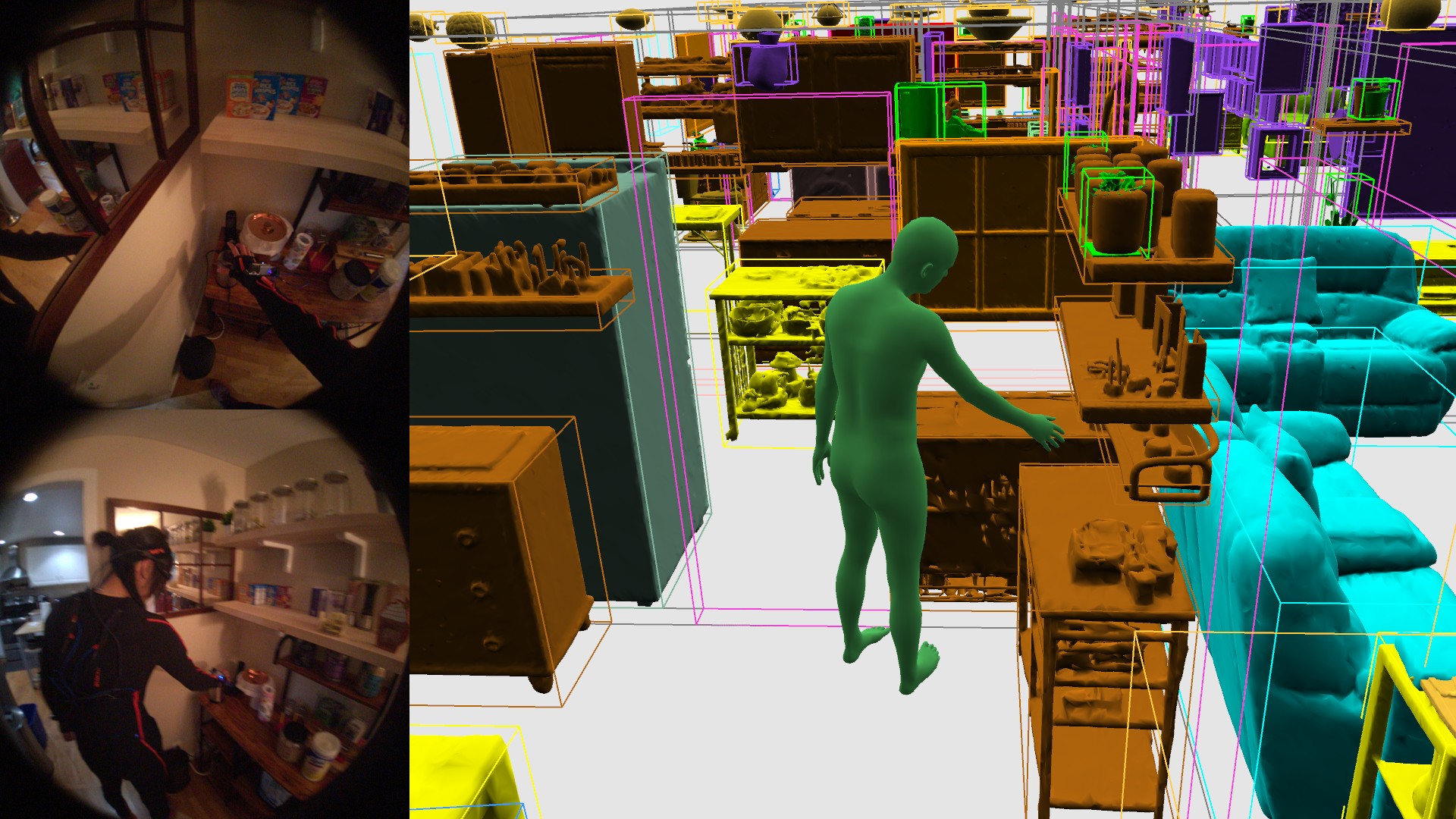}};
    \node(p12) at (p02.south) [anchor=north]{\includegraphics[width=\iw]{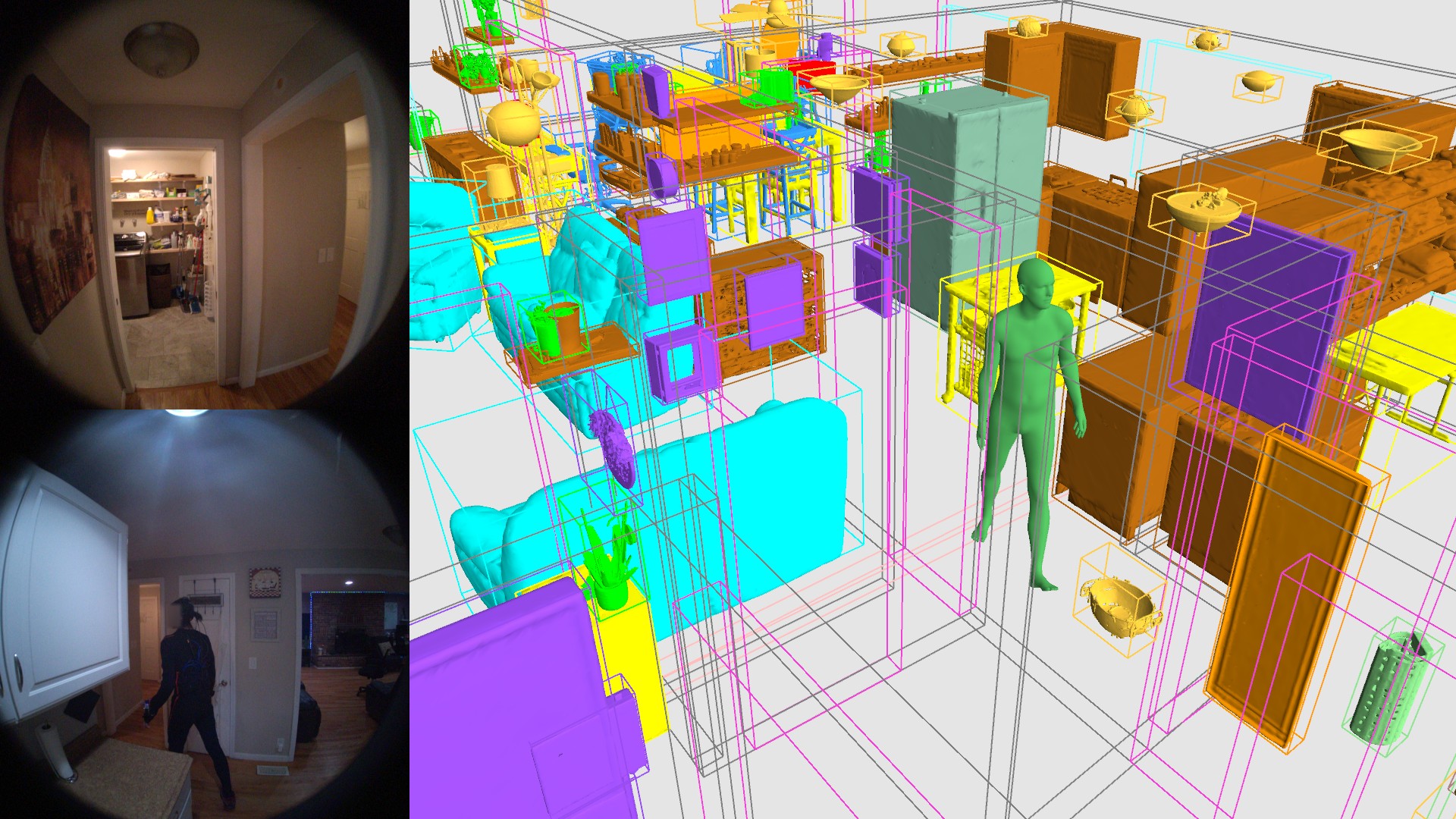}};
    \node(p13) at (p03.south) [anchor=north]{\includegraphics[width=\iw]{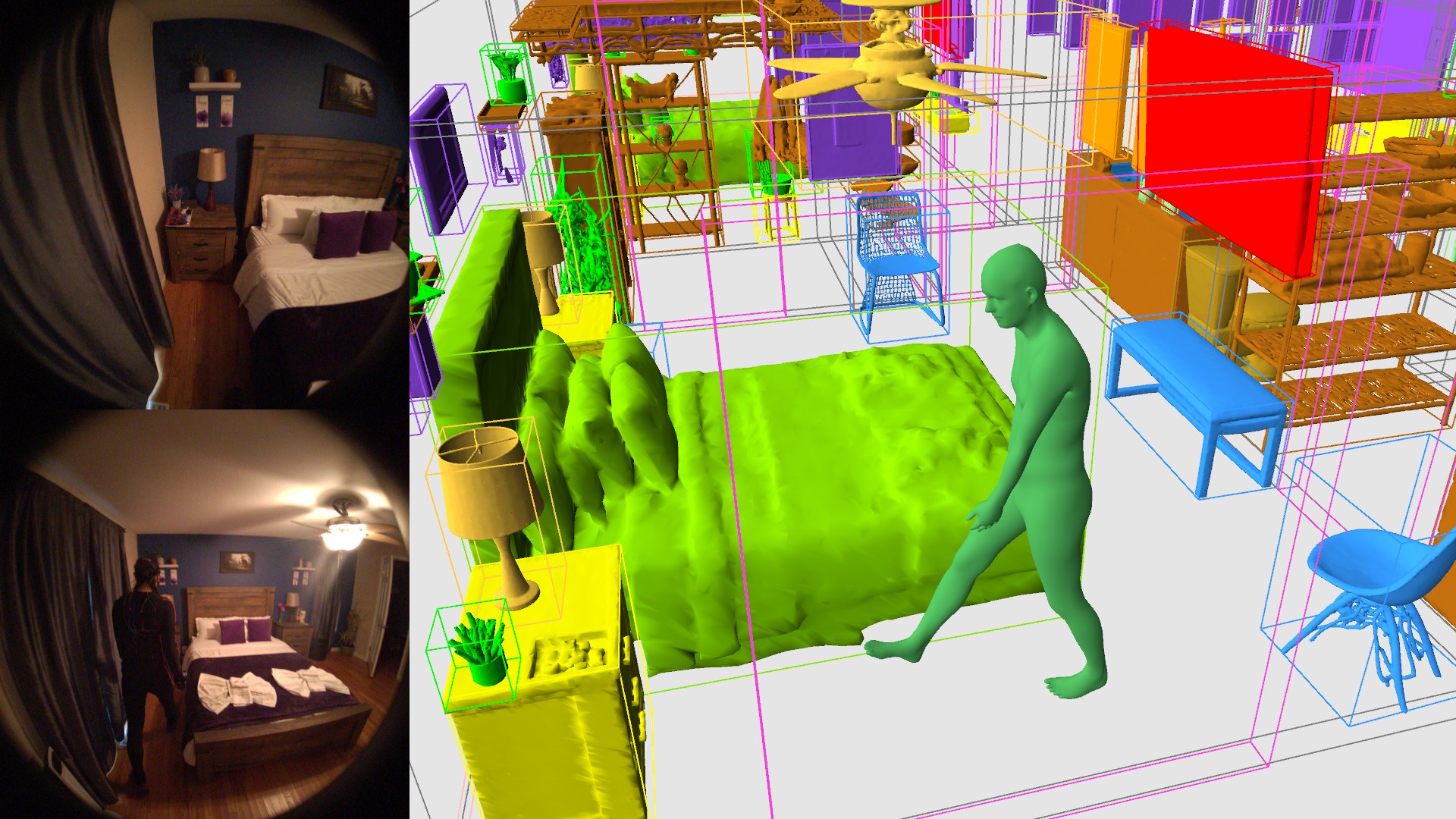}};

  \end{tikzpicture}

  The \nymeria dataset~\cite{nymeria24eccv}, released in 2024, is a large-scale collection of in-the-wild human activities captured with multiple egocentric wearable devices that are spatially localized and temporally synchronized.
  It provides body-motion ground truth recorded with a motion-capture suit, device trajectories, semi-dense 3D point clouds, and in-context narrations.
  In this paper, we upgrade \nymeria and introduce \nymeriapp.
  \nymeriapp features:
  (1) improved human motion in Momentum Human Rig (MHR)~\cite{mhr25} and SMPL~\cite{smpl15} formats;
  (2) dense 3D and 2D bounding box annotations for indoor objects and structural elements;
  (3) instance-level 3D object reconstructions; and
  (4) additional modalities \textit{e.g.,} basemap recordings, audio, and wristband videos.
  By consolidating these complementary modalities and annotations into a single, coherent benchmark, \nymeriapp strengthens \nymeria into a more powerful in-the-wild egocentric dataset.
  We expect \nymeriapp to bridge a key gap in existing egocentric resources and to support a broader range of research, including unique explorations of multimodal learning for embodied AI.
}

%% file: sections/1-intro.tex
\section{Introduction}\label{sec:intro}
Recent advances in artificial intelligence (AI) have increased demand for human-centric, context-aware algorithms and systems, particularly those enabled by egocentric wearables such as smart glasses and body-worn sensors.
These platforms capture multimodal signals that can provide rich context for foundation models, enabling seamless human-like understanding and interaction.
An important component of this paradigm is robust, context-aware human motion understanding.
Progress in this area importantly depends on large-scale datasets that combine diverse, high-quality motion with rich annotations beyond body pose, collected in natural, in-the-wild settings and spanning long-horizon activities.

The availability of egocentric datasets has grown rapidly in recent years.
However, most existing benchmarks lack either reliable ground-truth human motion or sufficiently rich context annotations~\cite{gtea18eccv, epickitchen18eccv, epickitchen20corr, trek22ijcv, egocom20pami,
  ego4d22cvpr, lamar22eccv, egotv23iccv, adt23iccv,
  egoschema23neurips, holoassist23iccv, ase24eccv, egovid5m24,
readingiwt25neurips, egolife25, lamaria25iccv, openego25, wearvox25, unihand226}.
At the same time, constructing human motion datasets involves persistent trade-offs among scale, diversity, sensing modalities, and annotation quality.
Although large-scale third-person motion datasets~\cite{motionx23neurips, motionxpp25, motionmillion25iccv} have made notable progress by pseudo-annotating Internet videos, these pipelines do not transfer well to egocentric viewpoints.
High-quality motion datasets collected in controlled environments (e.g., multi-view RGB(D) rigs or optical MoCap) provide detailed annotations but limit scene realism and the breadth of activities~\cite{egobody22eccv, egohuman23iccv, harmony4d24neurips, egoexo4d24cvpr, gorp25cvpr, embody3d25}.
Many motion datasets are also captured without egocentric wearables~\cite{amass19iccv, behave22cvpr, divatrack24cgf, parahome24, humoto25iccv, embody3d25}.
Moreover, despite the strong coupling between human motion and 3D environments, most motion datasets either omit object-level annotations or lack the geometric detail needed to model human--scene interaction.
A few exceptions incorporate scene context~\cite{unrealego22eccv, circle23cvpr, adt23iccv, bedlam23cvpr, trumans24cvpr, parahome24, hdepic25}, but the environments are often synthetic or restricted to a small number of carefully mapped scenes.
Conversely, many datasets provide detailed scene and object annotations~\cite{scannet17cvpr, scannet++23cvpr, replica19arxiv, ahmadyan2021objectron, lamar22eccv, egoobject23iccv, ase24eccv, mse25, objaverse23cvpr, objaversexl23neurips, dtc25cvpr} but focus on static environments without human interaction.
Existing human--object interaction datasets are frequently tailored to tabletop manipulation~\cite{grab20eccv, dexycb21cvpr, assembly10122cvpr, parahome24, hot3d25cvpr, gigahands25cvpr, humoto25iccv, egodex25, openego25}.
As a result, current datasets do not support a unified set of tasks spanning motion capture, egocentric multimodal perception, action understanding, and interaction modeling with rich scene context.
This gap has become a major bottleneck for advancing data-hungry models in human motion understanding, behavior analysis, and embodied AI.

Among existing efforts, the \nymeria dataset~\cite{nymeria24eccv} is an attractive resource.
Captured with Project Aria~\cite{aria23surreal} glasses and Aria-like wristbands, it records more than 300 hours of in-the-wild human activities and provides annotations including body motion, device trajectories, synchronization, semi-dense point clouds, and in-context narrations.
However, the released body motion is primarily optimized for spatial alignment with the glasses, without explicitly addressing artifacts introduced by inertial MoCap.
In addition, the dataset relies on a human model that is not widely adopted in academia, limiting interoperability with existing benchmarks.
Finally, the semi-dense point clouds are often insufficient for algorithms that require semantic and geometric scene context.
To address these limitations, we upgrade \nymeria and introduce \nymeriapp.
\nymeriapp substantially extends the original dataset along four dimensions.
First, it provides improved body motion annotations, increasing the accuracy and robustness of estimated body shape and pose.
We release results in two widely used, open-source parametric human models: Momentum Human Rig (MHR)~\cite{mhr25} and SMPL~\cite{smpl15,smpl23}.
Second, we add 3D and 2D bounding box annotations for all indoor recordings to support scene-aware motion understanding.
Using a closed-set taxonomy of 19 common categories, we label more than 10K unique object instances, spanning both movable objects (e.g., tables, chairs, storage units, and lamps) and structural elements (e.g., walls, floors, and stairs).
We further complement these annotations with an open-set taxonomy in selected scenes, labeling an additional 10K unique objects to capture richer semantic context.
Third, we introduce densely populated, spatially grounded 3D object reconstructions, facilitating fine-grained interaction-aware motion understanding and embodied reasoning.
Finally, \nymeriapp releases additional modalities excluded from prior releases, including dedicated scans of each indoor location that capture complete static scene geometry, as well as auxiliary sensor streams (e.g., wristband videos, headset audio, and calibration data).

By combining accurate 3D human motion with dense object annotations at scale, \nymeriapp bridges the gap between in-the-wild egocentric perception, rich scene context, and human motion that has constrained prior datasets.
We expect this release to enable a broad range of tasks, including environment-aware motion tracking and synthesis, action and interaction understanding, object detection, 3D reconstruction, and multimodal learning for embodied contextual AI.

%% file: sections/2-related-work.tex
\section{Related Works}\label{sect:recap}

\textbf{Egocentric Multimodal Datasets.}
Egocentric datasets have been developed for over a decade.
Early efforts such as EgoHands~\cite{egohands15iccv}, EGTEA~\cite{gtea18eccv}, EPIC-KITCHENS~\cite{epickitchen18eccv, epickitchen20corr}, EgoCom~\cite{egocom20pami}, H2O~\cite{h2o21cvpr}, DexYCB~\cite{dexycb21cvpr}, and Assembly101~\cite{assembly10122cvpr} primarily comprise small-scale collections recorded with head-mounted cameras or AR/VR headsets, and focus on hand--object interactions and activity understanding.
Ego4D~\cite{ego4d22cvpr} was the first large-scale in-the-wild egocentric dataset, capturing diverse daily activities using wearable glasses and leveraging human narrations as a key annotation signal.
More recently, the growing availability of AR/VR headsets~\cite{quest, hololens, vive, applevisionpro} and smart glasses~\cite{vuzix, aria23surreal, rayban} has led to a rapid expansion of multimodal egocentric datasets.
For instance, HoloAssist~\cite{holoassist23iccv} provides 166+ hours of paired performer--instructor recordings to support interactive assistant research.
EgoExo4D~\cite{egoexo4d24cvpr} captures 180+ hours across eight skill categories to study skilled activities.
Nymeria~\cite{nymeria24eccv} records 300+ hours of daily activities for understanding egocentric body motion.
EgoLife~\cite{egolife25} introduces a long-horizon question answering benchmark by collecting 266+ hours of footage from six people living in an apartment for a week.
LaMAria~\cite{lamaria25iccv} captures more than 22 hours of ground-truth device localization to benchmark SLAM.
DTC~\cite{dtc25cvpr} collects 2,000+ 3D object scans to benchmark object reconstruction.
EgoDex~\cite{egodex25} records 800+ hours of table-top hand--object manipulation.
Reading-in-the-Wild~\cite{readingiwt25neurips} records 100+ hours with ground-truth gaze for recognizing reading activities.
UniHand2.0~\cite{unihand226} collects 35K+ hours of egocentric human demonstrations for robot learning.
In addition to these large-scale efforts, numerous smaller datasets have been proposed~\cite{hps21cvpr, egobody22eccv, unrealego22eccv, egohuman23iccv, adt23iccv,
harmony4d24neurips, mmcsg24, hot3d25cvpr, egopressure25cvpr, egoppg25iccv, egomimic25icra}.
Despite this progress, there remains a lack of datasets that consolidate comprehensive annotations within a single corpus, which limits research on cross-modality reasoning.
\nymeriapp addresses this gap by augmenting the open-sourced \nymeria dataset with accurate parametric human motion, 3D object bounding boxes, and 3D object reconstruction, thereby enriching its existing multimodal sensing, multi-device localization, synchronization, and in-context narration.

\textbf{Human Motion Datasets.}
Human motion datasets have a long history, spanning both laboratory motion capture and in-the-wild recordings.
A prominent effort is AMASS~\cite{amass19iccv}, which aggregates numerous MoCap datasets~\cite{accad,
bmlhandball15, bmlmovi20, bmlrub02, cmumocap, dancedb, dfaust, eyesjapandataset, grab20eccv,
contactdb19cvpr, hdm05, human4d20, humaneva10ijcv, kit15icra, kit16, kit21, moyo23cvpr, mosh14tog,
mosh14tog, poseprior15cvpr, soma21iccv, sfu, tcdhand12, totalcapture17bmvc, wheelposer24}
into a unified SMPL-(X) representation~\cite{smpl15, smpl23, smplx19cvpr}, enabling large-scale analysis and learning.
While AMASS offers high-fidelity and diverse motion, it lacks environmental and interaction context.
Recent datasets therefore incorporate object motion and human--object interactions~\cite{adt23iccv, circle23cvpr, parahome24, trumans24cvpr, humoto25iccv}.
Beyond MoCap, RGB-(D) cameras are widely used to capture motion, either from monocular videos~\cite{physcap20tog, frankmocap21iccv, humanml3d22cvpr, smplerx23neurips, slahmr23cvpr, motionx23neurips, multihmr24eccv, motionmillion25iccv, satmhr25cvpr, prompthmr25cvpr} or multiview systems~\cite{panoptic15iccv, egobody22eccv, egohuman23iccv, harmony4d24neurips, egoexo4d24cvpr, humoto25iccv, embody3d25}.
For monocular capture, Motion-X~\cite{motionx23neurips, motionxpp25} and MotionMillion~\cite{motionmillion25iccv} are notable for their scale, leveraging large collections of internet video.
For multiview capture, EgoExo4D~\cite{egoexo4d24cvpr} and Embody3D~\cite{embody3d25} represent two large-scale efforts.
However, monocular approaches are poorly suited to egocentric viewpoints due to limited visibility of the body, and multiview systems typically trade off scene realism and activity coverage.
To enable in-the-wild motion capture, wearable MoCap suits have emerged as a practical alternative~\cite{totalcapture17bmvc, 3dpw18eccv, hps21cvpr, empose21iccv,
emdb23iccv, divatrack24cgf, mocapee24cvpr, nymeria24eccv}.
To mitigate artifacts inherent to non-vision-based tracking, prior work combines IMUs with vision~\cite{totalcapture17bmvc, 3dpw18eccv}, constrains motion locally~\cite{divatrack24cgf}, or optimizes motion using 3D scene context~\cite{hps21cvpr, mocapee24cvpr} or egocentric headset tracking~\cite{nymeria24eccv}.
Among these, \nymeria is a particularly large-scale in-the-wild egocentric collection, yet it still exhibits notable motion artifacts.
Only a subset of motion datasets explicitly targets egocentric capture~\cite{hps21cvpr, egobody22eccv, adt23iccv, egohuman23iccv, harmony4d24neurips, egoexo4d24cvpr, nymeria24eccv, hdepic25}.
In such settings, scene context is often represented either via dense reconstructions~\cite{egobody22eccv} or point clouds~\cite{hps21cvpr, egoexo4d24cvpr, nymeria24eccv}.
While ADT~\cite{adt23iccv} and HD-EPIC~\cite{hdepic25} provide semantic object context, their scene diversity remains limited.
Simulation can further scale data collection~\cite{unrealego22eccv, bedlam23cvpr, trumans24cvpr, circle23cvpr, egogen24cvpr}, but synthetic data often suffers from domain gaps due to imperfect noise and appearance models.

\textbf{Semantic Bounding Box Datasets.}
3D bounding box benchmarks vary widely in (i) the underlying sensing modality and data format, (ii) dataset scale, and (iii) the granularity and taxonomy of annotated object categories.
Traditional benchmarks include single-frame RGB-D datasets such as SUN RGB-D~\cite{song2015sun},
large-scale room-scanning datasets such as ScanNet~\cite{scannet17cvpr, scannet++23cvpr} and ARKitScenes~\cite{baruch2021arkitscenes},
and object-centric scanning datasets such as Objectron~\cite{ahmadyan2021objectron} and CO3D~\cite{reizenstein2021common}.
Omni3D~\cite{brazil2023omni3d} further unifies several of these sources (along with autonomous driving and simulation datasets) to train single-view 3D object detectors that generalize across domains.
With the emergence of egocentric video, several smaller-scale 3D bounding box datasets have also been introduced.
Notably, AEO~\cite{efm3d24} provides a diverse evaluation set for egocentric 3D object detection, while ADT~\cite{adt23iccv} offers 3D box annotations in a single-room environment.
The large performance gap observed when applying scanning-dataset-trained 3D detectors to AEO motivates the release of \nymeriapp.
In analogy to recent trends in 2D detection, a key direction in 3D detection is open-set, category-agnostic localization.
CA-1M~\cite{ca1m24cvpr} advances this goal by adding open-set 3D bounding boxes to ARKitScenes.
Complementarily, \nymeriapp provides 12k open-set 3D bounding boxes for training and five scenes with exhaustive 3D OBB annotations for validation.

\textbf{Object Reconstruction Datasets.} While numerous datasets have driven recent progress in 3D vision, they typically force a trade-off between detailed object geometry and natural environment context. Existing 3D datasets largely focus on isolated objects (e.g., ShapeNet \cite{chang2015shapenet}, Objaverse \cite{deitke2023objaverse,deitke2023objaversexl}, ABO \cite{collins2022abo}, GSO \cite{downs2022google}, DTU \cite{aanaes2016large}, uCO3D \cite{liu2025uncommon}), which inherently lack the realistic clutter, lighting, and occlusions of natural environments. Conversely, real-world scene datasets (e.g., ScanNet \cite{dai2017scannet}, ScanNet++ \cite{yeshwanth2023scannet++}, Matterport3D \cite{chang2017matterport3d}, Replica \cite{straub2019replica}) capture natural context but represent scenes as monolithic meshes, resulting in incomplete geometry, limited per-object resolution, and an inability to cleanly separate instances from their surroundings. Synthetic scene datasets (e.g., Hypersim \cite{roberts2021hypersim}, 3D-FRONT \cite{fu20213d}, SceneSmith \cite{pfaff2026scenesmith}, Aria Synthetic Environments \cite{avetisyan2024scenescript}) provide perfect instance separation but suffer from a well-documented sim-to-real gap due to simplistic layouts and an absence of natural physical clutter. Other approaches attempt to populate real scene scans by retrieving and aligning CAD assets (e.g., Scan2CAD \cite{avetisyan2019scan2cad}, ScanNotate \cite{rao2025leveraging}, Metascenes \cite{yu2025metascenes}, LiteReality \cite{huang2025literealitygraphicsready3dscene}); however, these CAD models serve only as approximate proxies and fail to capture the exact, fine-grained geometric shape of the actual objects. While a few datasets do capture exact real objects within real scenes, they are heavily constrained in scale: the ShapeR \cite{siddiqui2026shaper} evaluation set consists of only 7 sparsely annotated recordings, and Aria Digital Twin \cite{pan2023aria} is restricted to two scenes. In contrast, our dataset provides dense, instance-level real 3D geometry for objects situated naturally in real-world scenes across 47 distinct recordings, effectively bridging the gap between isolated object geometry and complex scene context.

%% file: sections/3-nymeria-plus-plus.tex
\section{\nymeriapp Dataset}
\label{sect:nymeriapp}

\subsection{Recap of Nymeria Dataset}
In this section, we briefly recap the \nymeria dataset~\cite{nymeria24eccv}.
\nymeria comprises more than 300 hours of in-the-wild human activities recorded from 256 participants across 50 indoor and outdoor locations and 20 unscripted scenarios.
Each participant wore Project Aria glasses~\cite{aria23surreal} and two Aria-like wristbands (miniAria), which record RGB, grayscale, and eye-tracking videos, IMU measurements, barometric pressure, and audio.
Participants also wore an XSens MoCap suit~\cite{xsens} that captures full-body kinematics at 240~Hz.
In each recording session, an observer wearing Project Aria glasses captured third-person views of the participant.
All devices are time-synchronized and localized in a shared metric 3D coordinate system.
The original release provides annotations including 6-DoF device trajectories, semi-dense point clouds, gaze, XSens kinematic motion, and retargeted parametric body motion produced using the Meta Momentum library~\cite{momentum}.
It also includes human narrations at three levels of granularity (motion narration, atomic actions, and activity summaries).
Since release, \nymeria has supported a range of research directions, including motion tracking and generation~\cite{hmd2253dv, egolm25cvpr}, activity understanding~\cite{egolm25cvpr, ego4o25cvpr, wacu25iccv}, video generation~\cite{peva25, egotwin25}, and dynamic reconstruction~\cite{4dgt25neurips}.

Despite its scale and rich multimodal sensing, the original \nymeria dataset has several limitations that motivate the extensions introduced in \nymeriapp. 
First, the provided human motion is obtained via a naïve optimization that aligns the XSens motion with the Aria world frame, but does not explicitly address motion artifacts.
In practice, the most prominent errors occur at the end effectors, manifesting as inaccurate wrist positions and noticeable foot sliding. 
Second, environmental context is represented as a semi-dense point cloud reconstructed via VIO/SLAM,
which lacks both semantic understanding and dense surface geometry.
This limits the ability to reason about human motion under scene constraints, where semantics and detailed surfaces are often essential.
In the remainder of this section, we summarize the key components of \nymeriapp: \cref{sect:bodymotion} describes our motion optimization pipeline, \cref{sect:bb} presents the 3D/2D bounding box annotations, \cref{sect:shaper} introduces object shape reconstructions, and \cref{sect:moredata} details the additional modalities and recordings released with the dataset.

\input{sections/3.1-motion}

\input{sections/3.2-boundingbox}
\input{sections/3.3-shapes}
\input{sections/3.4-additional-data}

%% file: sections/3.1-motion.tex
\subsection{Body Motion Optimization}
\label{sect:bodymotion}

We provide ground-truth 3D human motion for the entire dataset by formulating an optimization framework that leverages both XSens and Aria streams.
While \nymeria~\cite{nymeria24eccv} retargets XSens motion to an anatomically inspired human model to obtain 3D motion, we identify and address three key limitations of that approach.
First, it uses a proprietary human model, limiting compatibility with widely adopted parametric body representations.
Second, it retargets XSens motion to the body model using an artist-defined skeleton mapping, which can be brittle under XSens calibration errors or out-of-distribution body shapes (e.g., very short or tall subjects), yielding implausible bone lengths.
Third, it does not leverage Aria streams during optimization; instead, it aligns the XSens-retargeted motion to the metric Aria 3D world in a post-processing step.
This can introduce artifacts such as arm--torso self-penetration (due to lower end-effector accuracy in inertial MoCap), temporal jitter, and foot sliding (due to per-frame mapping errors from the XSens frame to the Aria world).
In the following sections, we address each of the points above in more detail by introducing the body model used in \nymeriapp (\cref{sec:body_models}) and the optimization framework combining XSens and Aria streams (\cref{sec:body_optimization}).
To ensure compatibility with other human motion datasets proposed in the literature, we also provide all \nymeriapp motions in the SMPL(-X) body model parameterization~\cite{smpl15, smplx19cvpr}, leveraging an ad-hoc retargeting approach (\cref{sec:smpl_retargeting}).

\subsubsection{MHR Human Model}
\label{sec:body_models}
We build our framework on the recently released Momentum Human Rig (MHR) model~\cite{mhr25}.
MHR is a parametric human body model defined as a function $M(\beta, \theta)$ that outputs a 3D mesh with $N$ vertices.
It takes as input shape parameters $\beta$ (coefficients of a linear identity space) and skeletal parameters $\theta \in \mathbb{R}^{204}$.
The parameter vector $\theta$ controls a 127-joint skeleton and can be decomposed into 136 pose parameters, which specify joint angles, and 68 identity parameters, which capture subject-specific bone lengths.
We denote these components as $\theta^P \in \mathbb{R}^{136}$ and $\theta^I \in \mathbb{R}^{68}$, respectively.
We further denote by $J(\theta) \in \mathbb{R}^{127 \times (4 \times 4)}$ the per-joint affine transformations induced by $\theta$.
Given $J(\theta)$, mesh vertices are computed via linear blend skinning (LBS).

Since XSens does not provide reliable body-shape information, we omit $\beta$ from our optimization.
We instead model per-frame pose using $\theta^P$ and per-subject skeletal identity (i.e., bone-length parameters) using $\theta^I$.
This choice leverages a key property of MHR: unlike SMPL(-X)~\cite{smpl15, smplx19cvpr}, it decouples the body surface from the underlying skeleton, enabling fine-grained control over bone-length proportions.

\subsubsection{Motion Optimization with MHR}
\label{sec:body_optimization}

As in \nymeria, we use XSens software~\cite{xsens} to preprocess skeletal motion and Project Aria MPS~\cite{aria23surreal} to localize the Aria glasses and miniAria wristbands in a shared metric 3D world.
In \nymeria, XSens motion is retargeted to the body model and then rigidly transformed from the XSens coordinate frame to the Aria world via hand--eye calibration~\cite{handeye17}, without leveraging miniAria tracking signals or environmental context during optimization.
In contrast, we (i) regress subject-specific bone lengths to obtain robust identity estimates, (ii) use these estimates to improve retargeting, and (iii) jointly optimize MHR against the XSens-retargeted motion and Aria trajectories in the metric 3D world.

\textbf{Bone length regression.} We observe that XSens skeletons can exhibit inaccurate body proportions.
To mitigate this issue, we train a regressor that takes the subject-reported height as input and predicts MHR identity parameters $\theta^I$.
We use ridge regression with leave-one-out cross-validation and leverage the training set introduced in~\cite{mhr25}, which contains 7,110 scans registered to the MHR model. Although the regressor does not account for small height discrepancies due to footwear or hairstyle, we found it to be robust in practice.

\textbf{XSens motion retargeting.} We obtain an initial motion estimate by retargeting XSens motion to MHR.
Our method solves an inverse-kinematics optimization similar to~\cite{nymeria24eccv}: given 23 XSens skeleton segments and 79 anatomical landmarks, along with an artist-designed mapping to the MHR skeleton, we estimate per-frame pose parameters $\theta^P$ and per-subject identity parameters $\theta^I$.
During optimization, we strongly regularize $\theta^I$ toward the identity predicted by our regressor.
While the resulting motion sequence $\Theta^{X}$ is generally plausible, it is computed in the dead-reckoned XSens trajectory and therefore suffers from drift.
Moreover, as is common in inertial motion capture, errors accumulate along the kinematic chain, leading to wrist pose inaccuracies and occasional self-penetrations between the arms and torso.

\textbf{Joint XSens--Aria optimization.} To reduce drift and kinematic inaccuracies from the previous stage, we leverage Aria and wristband 6-DoF trajectories and formulate a joint XSens--Aria optimization framework.
This setting introduces several practical challenges.
First, XSens and Aria trajectories lie in distinct coordinate frames (sharing only the gravity direction); XSens global trajectory drifts over time, making the transformation aligning XSens and Aria world coordinates unknown and varying over time. Residual misalignment can then introduce inconsistencies between global and local motion, for example in the form of foot sliding.
Second, the placement of the Aria glasses and wristbands on the subject is only known up to calibration error and may vary slightly over time.
Third, we cannot rely on accurate foot--ground contact information: floor annotations (\cref{sect:bb}) are accurate only up to 
a 1-2 centimeter for indoor scenes and are unavailable outdoors; while XSens provides contact labels, we find them unreliable especially when subjects climb up stairs and transition across floors.

Our optimization framework takes as input retargeted motion $\Theta^{X}$ and Aria head trajectories $T^H$ plus wristband trajectories $T^{Wr}$ over a sequence. It outputs motion $\Theta^{W}$ in Aria world coordinates.
First, we seek a sequence of frame-varying transformations roughly aligning XSens and Aria coordinate frames.
As in~\cite{nymeria24eccv}, we assume a constant transformation $T_{X \rightarrow D}$ from the XSens head segment to the Aria device D and estimate its value via HandEye calibration~\cite{handeye17}.
Combining $T_{X \rightarrow D}$ with $T^H$, we obtain a sequence of per-frame rigid transformations converting motion $\Theta^{X}$ into Aria world, $\Theta^{W,rigid}$. Note that this is exactly the post-processing step used in~\cite{nymeria24eccv}.
While the approach can provide a preliminary alignment, it is not robust to inaccuracies in $T_{X \rightarrow D}$, in $T^H$, and in $\Theta^{X}$ (especially around the neck area). $\Theta^{W,rigid}$ exhibits temporal jitter and pose inaccuracies accumulating along the kinematic chain, particularly visible as foot sliding artifacts.

Taking $\Theta^{W,rigid}$ as initialization, we define an optimization problem seeking a refined motion $\Theta^W$ in the Aria coordinate frame. Namely, we minimize the following objective function:
\begin{equation}
  \label{eq:loss}
  \begin{split}
    \mathcal{L}(\Theta^W) = {} & \mathcal{L}_{T}(\Theta^W; T^H, T^{Wr})  +  \lambda_{X} \mathcal{L}_{X}(\Theta^W; \Theta^{X}) + \lambda_{l} \mathcal{L}_{l}(\Theta^W) + \lambda_{smooth} \mathcal{L}_{smooth}(\Theta^W_{root})  \\ & + \lambda_{g}\mathcal{L}_{g}(\Theta^W; \Theta^{X})  +  \lambda_{f}(w_{vel}) \mathcal{L}_{f}(\Theta^W; \Theta^{X})
  \end{split}
\end{equation}
where $\mathcal{L}_{T}$ is a robust loss penalizing position errors between MHR head and wrist joints and Aria trajectories, $\mathcal{L}_{X}$ enforces local joint angles to be similar to the ones retargeted from XSens, $\mathcal{L}_{l}$ enforces pre-defined limits on model parameters, $\mathcal{L}_{smooth}$ enforces temporal smoothness on the body root global translation and rotation, and $\mathcal{L}_{g}$ encourages similar body root joint orientation along the gravity direction between $\Theta^W$ and $\Theta^{X}$. To mitigate foot sliding, $\mathcal{L}_{f}$ enforces constant position for MHR foot joints in frames where feet are deemed in contact (when XSens foot contact labels are available) or where XSens foot velocities $vel_{X}$ are close to $0$ (velocity norms computed from $\Theta^X$, when contact labels are not available). To avoid motion discontinuities, for $\mathcal{L}_{f}$ we adopt an adaptive weighting schema defining $\lambda_{f}(w_{vel}) = \lambda_0 + \lambda_\alpha e^{-vel_{X}/v_\sigma}$, where $\lambda_0$, $\lambda_\alpha$, $v_\sigma$ are fixed hyperparameters; for frames in which feet are in contact, we set $vel_{X}$ to $0$.
The other $\lambda$ values in \cref{eq:loss} weight the contributions of the different error terms.

We minimize the objective in~\cref{eq:loss} sequentially, over batches of $2000$ frames, with an efficient Gauss-Newton solver, using the python API from the Momentum library~\cite{momentum}.

\subsubsection{MHR to SMPL Conversion}
\label{sec:smpl_retargeting}

To maximize interoperability with existing datasets and models, we additionally release all ground-truth motions in the SMPL(-X)~\cite{smpl15, smplx19cvpr} parameterization. We perform this conversion using the open-source SMPL-to-MHR tool released with MHR~\cite{mhr25}. For completeness, we briefly outline the procedure.

We first define an artist-curated correspondence between MHR and SMPL templates.
Concretely, we warp the SMPL template to match the MHR template geometry and associate each SMPL vertex with a surface point on the MHR template via triangle identifier and barycentric coordinates.
This correspondence enables converting an MHR mesh to the SMPL topology via barycentric interpolation.

Given an MHR motion sequence, we re-topologize each resulting MHR mesh to the SMPL topology and then solve an optimization problem to recover the corresponding SMPL parameters. Specifically, given a sequence of MHR meshes $M^\text{MHR}$ (re-topologized to SMPL), we optimize for the SMPL parameters $\Psi$ (including shape and pose) by minimizing the following objective:
\begin{equation}
  \label{eq:opt}
  \mathcal{L}(\Psi) = \mathcal{L}_{V}(\Psi; M^\text{MHR}) + \lambda_{E}\mathcal{L}_{E}(\Psi; M^\text{MHR})
\end{equation}
where ${L}_{V}$ penalizes the Euclidean distance between optimized SMPL vertices $M(\Psi)$ and target (re-topologized) MHR vertices and $\mathcal{L}_{E} = \sum_{(v, v') \in \varepsilon}||(M(\Psi)_v - M(\Psi)_{v'}) - (M^\text{MHR}_v - M^\text{MHR}_{v'})||$ penalizes edge discrepancies, over the set of SMPL template edges $\varepsilon$, with an $L1$ loss. $\lambda_{E}$ balances the contribution of the two terms.

Given the problem is highly non-convex, optimization proceeds in multiple stages to avoid getting stuck in local minima.
First, we optimize a set of identity parameters over a set of sparse frames, selected based on a given heuristic (e.g., pose diversity or geometric similarity with the rest template).
Then, we optimize for body pose and refine identity by minimizing the loss in \cref{eq:opt} according to a hierarchical scheme, fitting first global orientation and then all parameters simultaneously.
At each stage, we rely on the Adam optimizer provided by the pytorch library~\cite{pytorch}.

\begin{figure}[!]
  \centering
  \begin{tikzpicture}
    \node(p0){\includegraphics[width=0.99\textwidth]{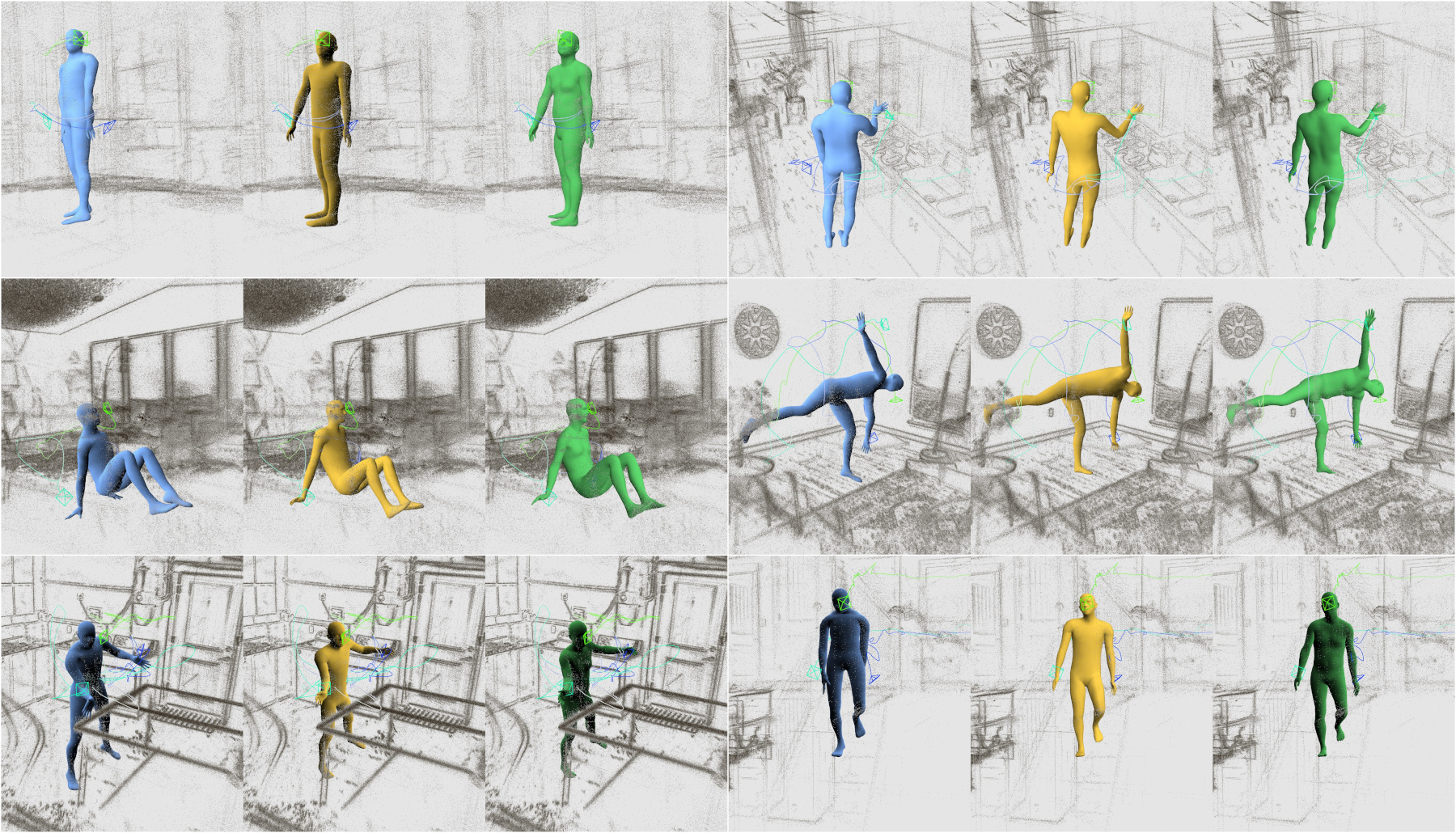}};
  \end{tikzpicture}
  \caption{\textbf{\nymeria and \nymeriapp motion samples}. We compare \nymeria results (blue) against the new MHR (yellow) and SMPL (green) ones. Each frame also shows Aria head and wrist trajectories. \nymeriapp results exhibit more realistic body proportions and better matching of wrist trajectories, resulting in fewer body self-penetrations and improved placement with respect to the 3D scene.}
  \label{fig:motion_comparison}
\end{figure}

\subsubsection{Metrics and Evaluations}

We evaluate the quality of the MHR ground-truth motions by comparing them against the original \nymeria ones. Based on metrics results, we identify and filter out $17$ problematic sequences from the original dataset -- mostly due to wrist recording issues. In total, \nymeriapp collects $1083$ sequences.

We consider three metrics:
\begin{itemize}
  \item Wrist distance error: We compute the distance (in cm) between the body model wrist joints and the miniAria wristband positions, per frame. Note that this error includes a constant offset (from the wrist joint to the wristband coordinate frame origin), which is however the same across models and thus ensures a fair comparison.
  \item Body self-penetration error: We leverage the collision error term implemented in the Momentum library~\cite{momentum}. It approximates the body as a set of geometric primitives and detects their collision, by iterating over pairs of primitives and checking if they overlap; overlap is determined by comparing their positions, directions, and radii. This collision detection routine is available for both MHR and the body model used in \nymeria.
  \item Foot sliding error: We consider the frames in which XSens detects foot contact, and compute the body model foot joint velocities in these frames. Following previous work~\cite{zhang2024rohm}, we consider foot sliding occurs if joint velocity exceeds 10cm/s during contact; we then compute the percentage of (contact) frames in which sliding is detected. We limit the computation to the heel joint since we find its contact more reliably detected by the XSens software.
\end{itemize}
For each metric, we compute an average error over all the \nymeriapp sequences after downsampling them to 30 fps. We also show visual results in \cref{fig:motion_comparison}, comparing \nymeria results (blue) versus the new motions in MHR (yellow) and SMPL (green) format.

As for wrist distance, the average error computed on the original \nymeria dataset is $14.32$cm, which decreases 
to $5.07$cm for \nymeriapp (please recall that this error also includes the offset between miniAria and wrist joint). Numbers are well reflected by visual results, showing how the body model wrists now better match miniAria trajectories.
A similar trend is confirmed by the body self-penetration error, whose average is $18.67$ in \nymeria and $2.44$ in \nymeriapp. 
In particular, self-penetration between arms and torso was problematic in \nymeria given its reliance on XSens motions only. By leveraging both XSens and Aria trajectories at optimization time, we are able to significantly reduce these artifacts in \nymeriapp.

Mitigating foot sliding proved challenging. The difficulty lies in the absence of reliable foot contact labels and of strong input signals for the lower-body motion (in Aria world, we can just rely on head and wrist streams, leaving the problem of reconstructing lower-body motion less constrained).
Nonetheless, we show improvement going from $35\%$ frames deemed as ``exhibit foot sliding'' in \nymeria down to $9.81\%$ in \nymeriapp.

%% file: sections/3.2-boundingbox.tex
\subsection{Bounding Box Annotations}
\label{sect:bb}

We annotate all \nymeriapp recordings with 3D oriented bounding boxes (OBBs) and their corresponding 2D projections to facilitate object-level 3D grounding and scene-aware motion reasoning.
We describe both our fixed-taxonomy and open-set 3D OBB annotation pipelines.
At a high level, we first collect dedicated ``basemap'' recordings for each indoor venue and annotate them with 3D boxes.
We then transfer these basemap annotations to each Nymeria recording by exploiting the shared coordinate system provided by localization.
Since objects may move across sessions, annotators subsequently verify and refine the transferred boxes and add sparse open-set OBB annotations.
Finally, we exhaustively annotate five basemaps using an open-set taxonomy to enable benchmarking of open-set 3D OBB detectors.
We provide detailed statistics and qualitative visualizations for the resulting 3D object annotations in the \nymeriapp release.

\subsubsection{3D Annotation Tool}
\begin{figure}[!]
\centering
\begin{tikzpicture}
\node(p0){\includegraphics[width=0.95\textwidth]{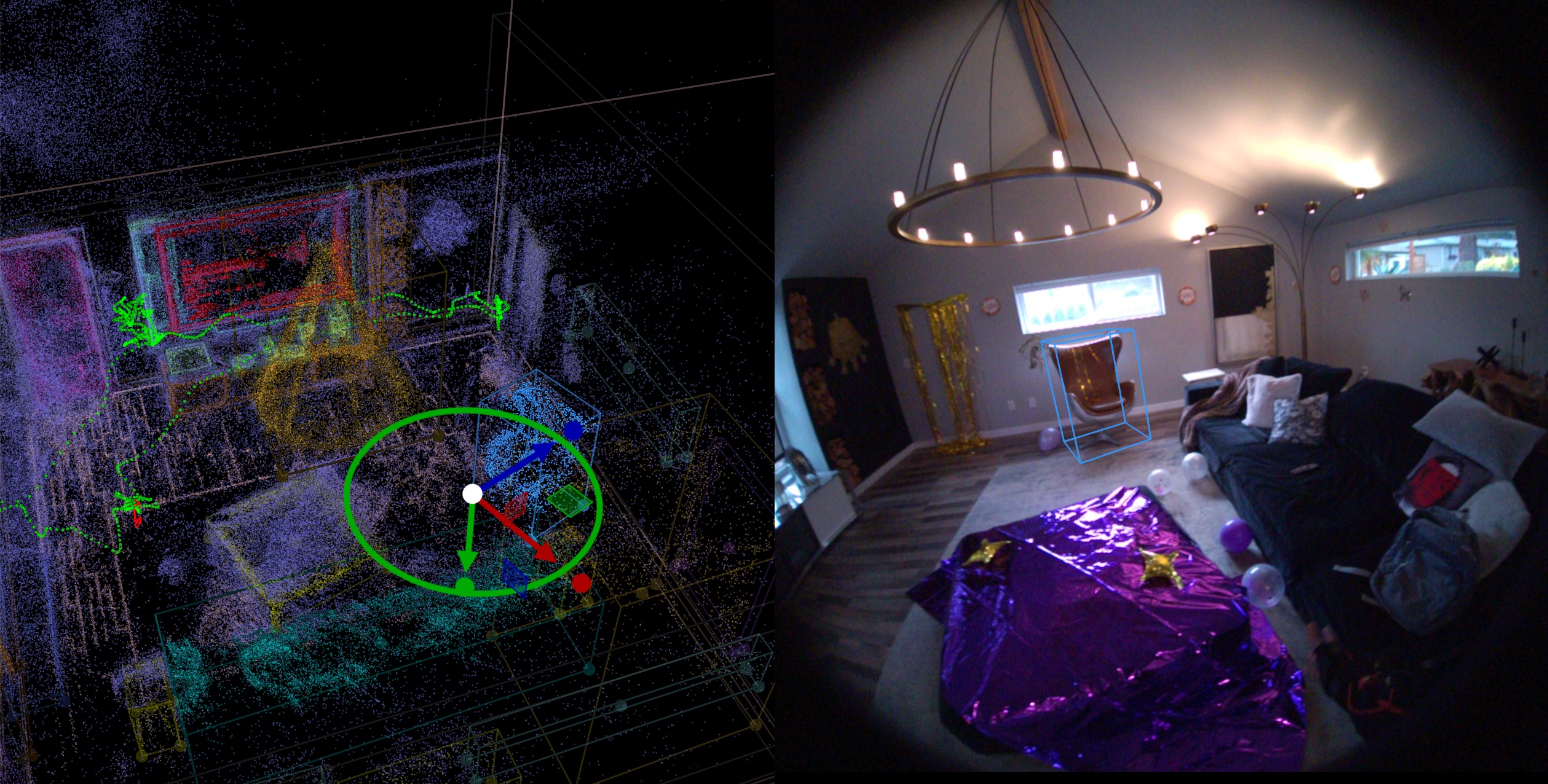}};
\end{tikzpicture}
\caption{\textbf{Boxy 3D OBB annotation tool interface.} (Left) The 3D point cloud and a 3D gizmo are used to modify all nine degrees of freedom of the box directly in 3D. (Right) The current 3D box annotation is projected into the RGB image. Annotators can iterate across views to inspect alignment.}
\label{fig:boxy}
\end{figure}

We developed a 3D OBB annotation tool, Boxy (\cref{fig:boxy}).
Boxy takes as input the original video, online-estimated camera intrinsics, optimized 6-DoF device trajectories, and the semi-dense point cloud.
Annotators place 3D OBBs around static objects and adjust their 9-DoF directly in 3D, using both the point cloud and the projected box overlays in the image for guidance and verification. To facilitate annotation, Boxy can automatically select informative multi-view frames based on the observability of the semi-dense point cloud.

\begin{figure}[!]
\centering
\begin{tikzpicture}
\node(p0){\includegraphics[width=0.98\textwidth]{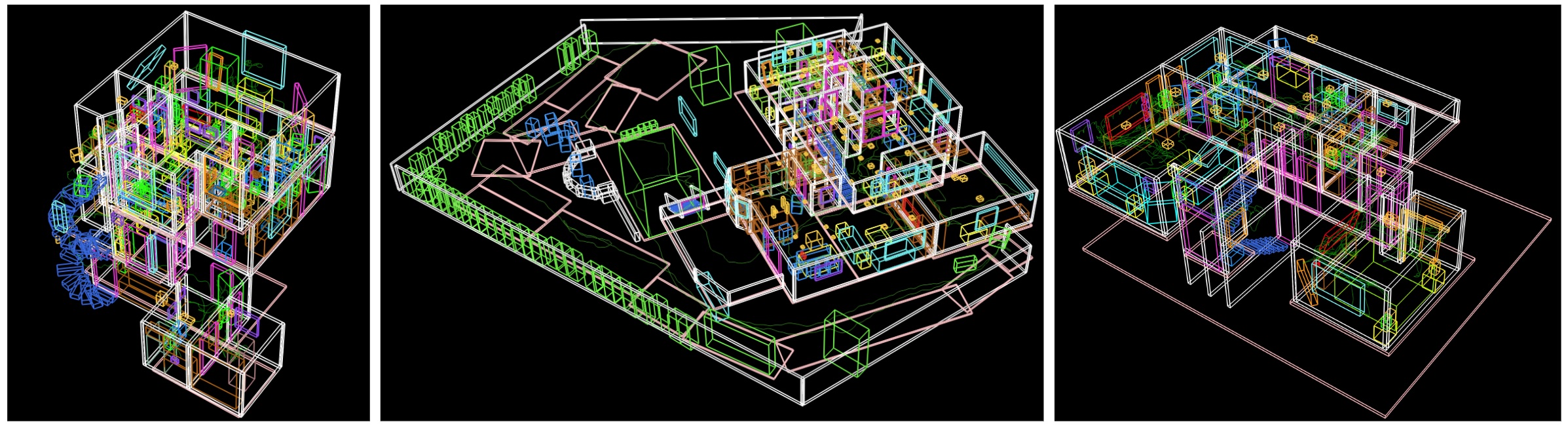}};
\end{tikzpicture}
\caption{\textbf{Basemap 3D OBB annotations}. We visualize basemap annotations for three venues from \nymeriapp.}
\label{fig:basemaps3d}
\end{figure}

\begin{figure}[!]
\centering
\begin{tikzpicture}
\node(p0){\includegraphics[width=0.98\textwidth]{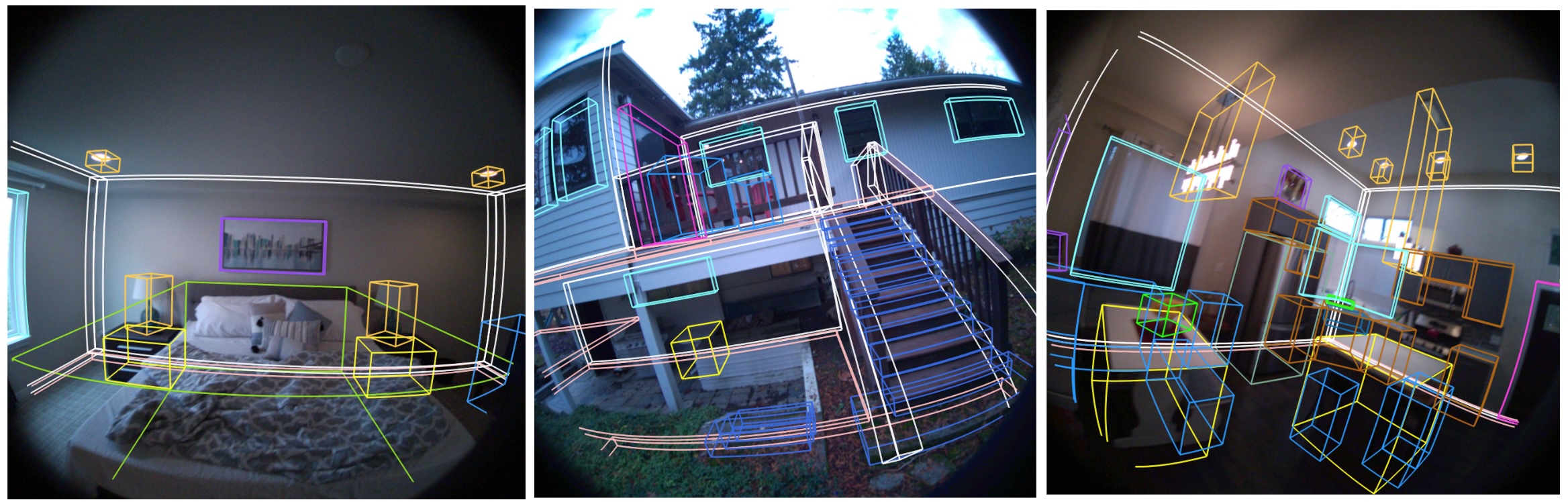}};
\end{tikzpicture}
\caption{\textbf{Basemap 2D OBB annotations}. Visible 3D OBBs from three venues are rendered into sample RGB frames.}
\label{fig:basemaps2d}
\end{figure}

\subsubsection{2D Bounding Box}
For 3D OBB detection -- both single-view methods such as Cube R-CNN~\cite{brazil2023omni3d} and multi-view methods such as EVL~\cite{efm3d24} -- it is important to know which 3D boxes are visible from a given viewpoint and the corresponding 2D bounding boxes (2DBBs) they induce.
Manually annotating per-frame 2DBBs and visibility is prohibitively expensive due to the large number of views per object; we therefore compute these quantities automatically.
To estimate visibility, we require that (i) the 3D OBB projects into the image, (ii) at least two semi-dense points fall inside the 3D box, and (iii) at least 85\% of the projected 3D box lies within the valid image region.
We compute the associated 2DBB by projecting the 3D box into the image and taking the axis-aligned bounding box enclosing samples along its edges.
This automated procedure is not error-free and can introduce occasional temporal flicker when visualizing projected boxes.
Nevertheless, we found the resulting annotations to be sufficient for training and evaluating 3D OBB detectors, and we leave further temporal smoothing and refinement to future work.

\subsubsection{Basemaps and 3D Annotation Transfer}
\textbf{Basemaps.} \nymeria contains 1,100 sequences recorded across 50 locations, including 47 indoor house venues.
Annotating every recording independently would require substantial manual effort.
We therefore exploit the fact that \nymeria localizes all recordings captured at the same venue within a shared coordinate system.
Since recordings at a given venue are always collected within 2--4 consecutive days, the environment tends to remain largely static.
During \nymeria data collection, most indoor venues were scanned using a single pair of Project Aria glasses to create dedicated ``\emph{Basemap}'' recordings. 
These basemaps are designed to observe the full 3D space from diverse viewpoints, 
where operators were instructed not to modify or interact with the scene to preserve its static layout.
When a venue does not include a basemap, we manually select one of the existing recordings that covers the space well while exhibiting minimal human--scene interaction (five basemaps from \emph{Welcome To My Home} and three from \emph{Scavenger Hunt}).
In total, we collected 47 basemap recordings, one per indoor house venue, each typically 5--15 minutes long.
\Cref{fig:basemaps3d} visualizes 3D OBB annotations for three venues, and \cref{fig:basemaps2d} shows their projections onto RGB frames.

\textbf{Annotation transfer and refinement.} After annotating the basemaps, we transfer their boxes to all recordings captured at the same location using the shared coordinate frame, and then refine each 3D OBB.
Because most objects are static (\textit{e.g.,} walls, floors, cabinets, and dressers), this refinement is relatively lightweight.
Annotators remove boxes for objects that are no longer present or not observed in the target recording, and adjust boxes for objects that have been slightly moved.
We note that some recordings may capture previously unseen rooms or areas; in such cases, newly observed objects may not have corresponding annotations.

\subsubsection{Fixed Set Taxonomy}
We adopt 19 commonly occurring object categories to define a closed-set taxonomy for 3D OBBs.
This taxonomy is similar to the classes used in EFM3D~\cite{efm3d24} (see \cref{app:fixed-taxonomy} for detailed definitions).
In addition to common objects (e.g., beds, chairs, lamps, and tables), our taxonomy includes structural classes such as floors, walls, and stairs, which are informative for scene-aware motion reasoning.

\begin{figure}[t]
\centering
\begin{minipage}[t]{0.65\textwidth}
\vspace{0pt}
\centering
\includegraphics[width=\linewidth]{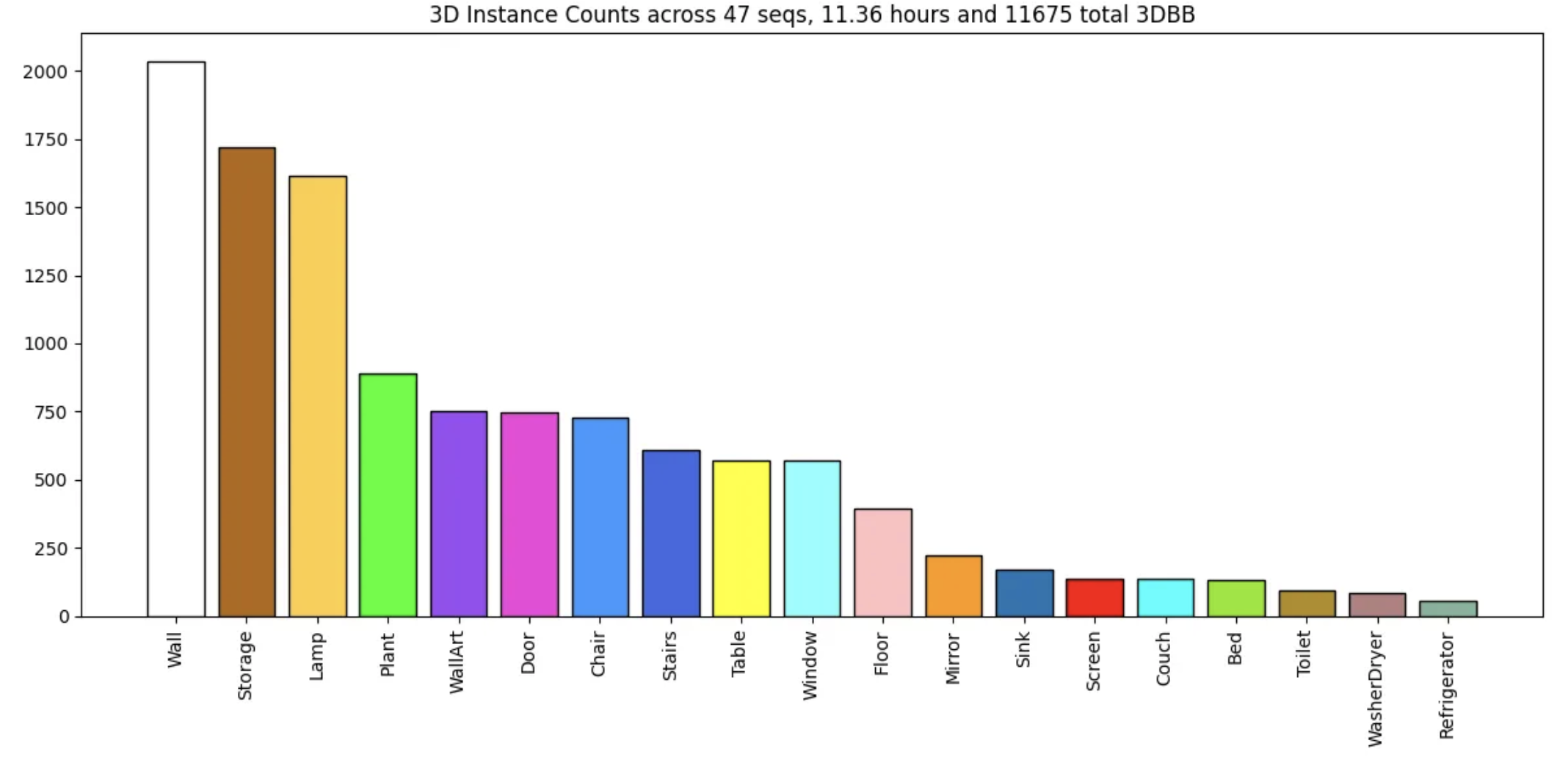}
\end{minipage}\hfill
\begin{minipage}[t]{0.32\textwidth}
\vspace{0pt}
\caption{\textbf{Basemap unique 3D objects histogram}. We count the number of unique 3D OBBs across the basemaps for each class. Since we transfer these OBBs to the individual scenario recordings, this is the fundamental distribution over observed objects. Note that transferring the OBBs to scenario recordings increases the viewpoint diversity substantially over just the basemap sequences.}
\label{fig:3dbb_statistics}
\end{minipage}
\end{figure}

\begin{figure}[!]
\centering
\begin{tikzpicture}
\node(p0){\includegraphics[width=0.95\textwidth]{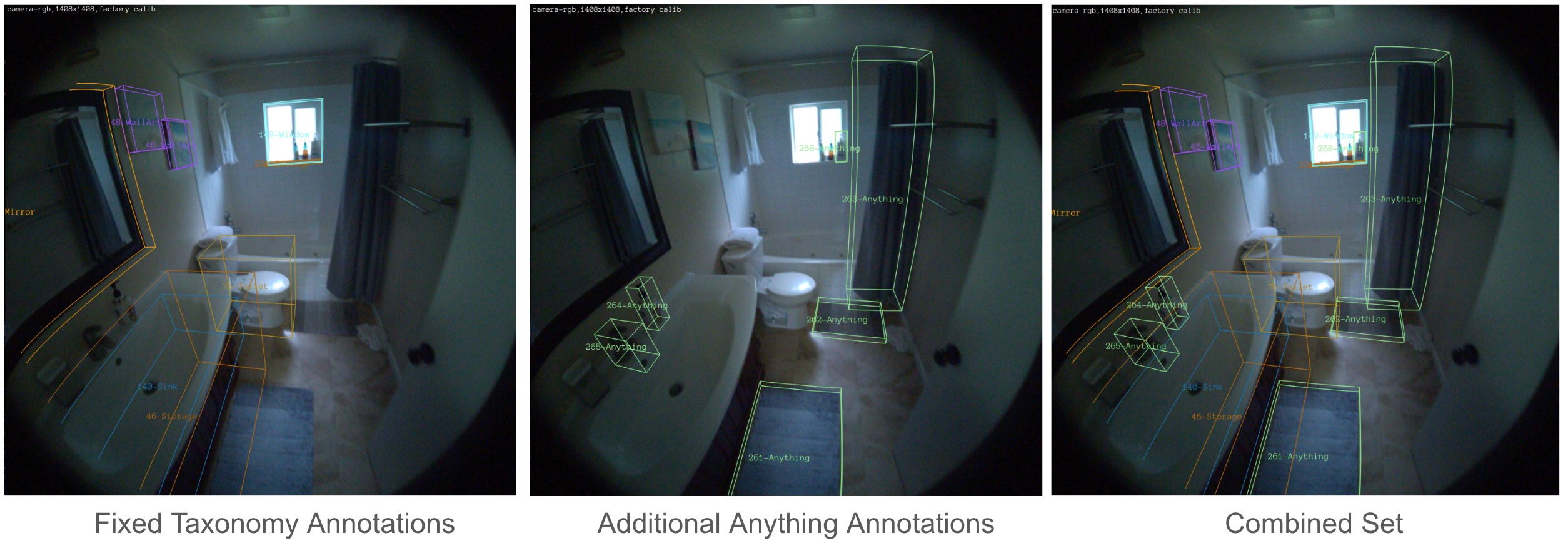}};
\end{tikzpicture}
\caption{\textbf{Example sparse open-world annotations}.  (left) A fixed taxonomy of objects and structure are first annotated. (middle) Additional "anything" objects are added, roughly 12 per recording. (right) All objects visualized together.}
\label{fig:anything_example}
\end{figure}

\subsubsection{Open Set Taxonomy}

\textbf{Sparse open set annotations.}
We further augment each recording with a set of \textit{Anything} objects, capturing instances that fall outside the fixed taxonomy.
These annotations often correspond to objects that are frequently moved (e.g., cups, bags, and bottles).
Because Boxy does not currently support dynamic-object annotation, we do not label \textit{Anything} objects on basemaps; instead, we annotate a sparse set independently for each recording.
On average, we annotate 12 \textit{Anything} objects per recording.
As this process is intentionally non-exhaustive due to annotation cost, the resulting labels provide broad coverage of long-tail objects without requiring complete scene annotation. 
Overall, we collect approximately \textit{12K Anything} object annotations across \nymeriapp, which is comparable in scale to the fixed-taxonomy 3D OBB annotations.
\Cref{fig:anything_example} shows a qualitative example.

\begin{figure}[!]
\centering
\begin{tikzpicture}
\node(p0){\includegraphics[width=0.95\textwidth]{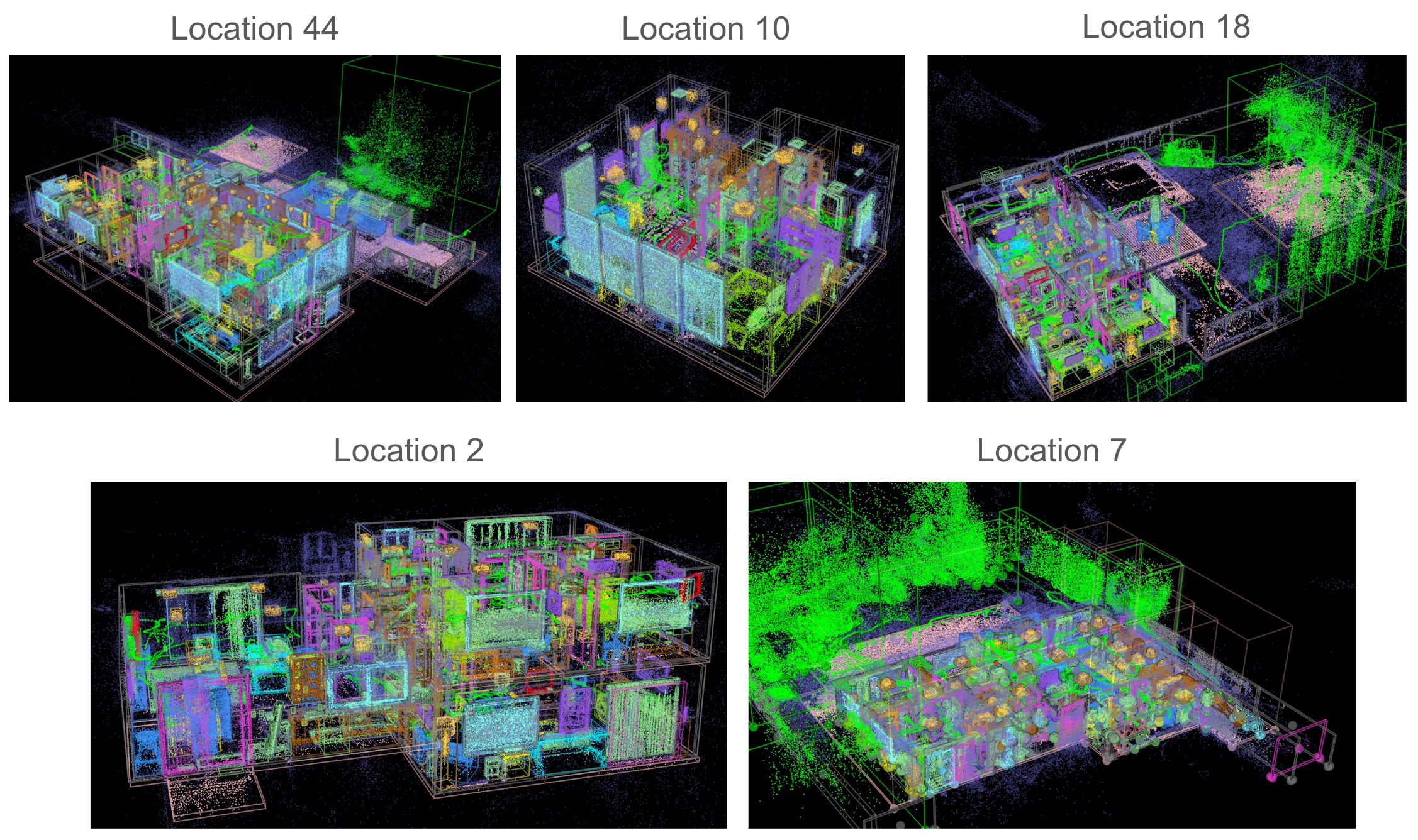}};
\end{tikzpicture}
\caption{\textbf{Densely annotated venues.} We annotated five static basemaps densely with "every" object annotated.}
\label{fig:dense_example}
\end{figure}

\begin{figure}[!]
\centering
\begin{tikzpicture}
\node(p0){\includegraphics[width=0.95\textwidth]{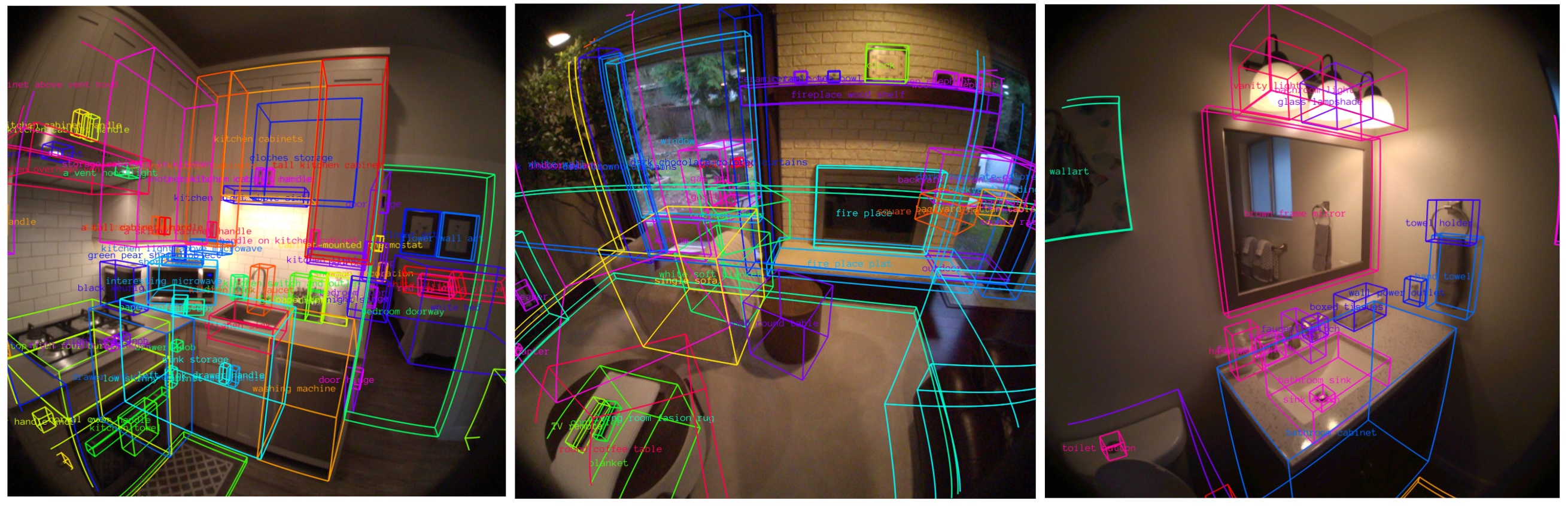}};
\end{tikzpicture}
\caption{\textbf{Example dense annotations.} Examples of dense, open-world annotation are shown with the corresponding text caption.}
\label{fig:open_rgb}
\end{figure}

\textbf{Dense open set with captions.}
For five basemaps, we provide exhaustive open-set annotations to serve as an evaluation set for open-world 3D OBB detection (\cref{fig:dense_example}).
While the notion of an ``object'' can be subjective, we aim to annotate the scenes as densely as possible.
We prioritize objects that are movable or that correspond to interactive parts (\textit{e.g.,} handles, knobs, and switches), and we omit fine-scale textures and highly non-cuboidal structures (e.g., cables), as well as low-level structural elements such as individual floor tiles.
Across the five densely annotated venues, we label a total of 2,896 objects (579 per venue on average).
In addition, annotators provide a short descriptive caption for each object (adjective + noun) to facilitate identification within the scene (e.g., ``cabinet handle'' or ``red velvet curtain''). \Cref{fig:open_rgb} shows examples of these dense annotations with captions.

%% file: sections/3.3-shapes.tex
\subsection{Object Reconstruction with ShapeR}
\label{sect:shaper}

\begin{figure}[!]
\centering
\begin{tikzpicture}
\node(p0){\includegraphics[width=0.99\textwidth]{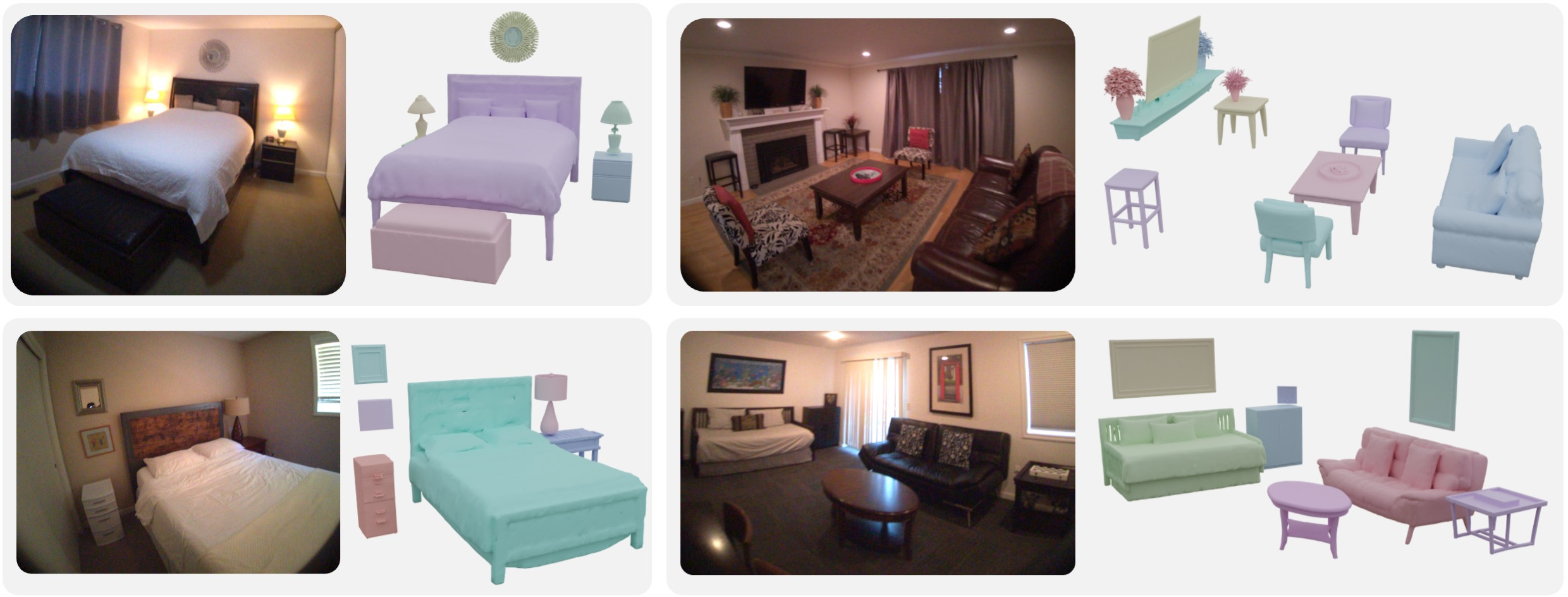}};
\end{tikzpicture}
\caption{\textbf{NymeriaPlus Shape Annotations.} Examples of recordings with shape annotations. Different colors represent distinct shape instances.}
\label{fig:shaper_examples}
\end{figure}

To provide a more detailed geometric representation beyond coarse spatial bounds, we also release dense metric 3D shape reconstructions inferred via ShapeR~\cite{siddiqui2026shaper} for a subset of the 3D bounding boxes (Figure~\ref{fig:shaper_examples}). Specifically, we exclude basic structural elements (e.g., walls, floors, ceilings, doors, stairs, and windows) whose geometry is already well-approximated by their bounding volumes. For the remaining object categories, we leverage the probabilistic nature of ShapeR to generate four candidate meshes per instance. To ensure high geometric fidelity, these proposals undergo a rigorous manual quality control process. Human annotators evaluate and rate the candidate reconstructions via a dedicated web interface. In the final dataset, we release the single highest-rated mesh for each bounding box, provided it meets a quality threshold. In the remainder of this section, we detail the inference procedure used to generate these candidates and outline our manual quality-control protocol.

\subsubsection{Candidate Reconstruction Generation}

ShapeR~\cite{siddiqui2026shaper} requires the following instance-specific inputs for reconstruction: a semi-dense point cloud, multiple posed images, and a text description. We process the NymeriaPlus sequences to extract these inputs for each 3D bounding box. Because a bounding box encompasses all spatial geometry within its limits, a naive point extraction often includes background clutter or parts of adjacent objects (e.g., table points captured within a chair's bounding box). To isolate the target geometry, we project the enclosed 3D points into the 2D camera frames and apply SAM2 \cite{ravi2024sam} masks to filter out non-instance points, resulting in a clean, object-centric point cloud.

Using the visibility associations between these filtered 3D instance points and the 2D frames, we identify all images where the object is visible. From this set, we sample 16 frames using a greedy view-selection heuristic designed to maximize viewpoint diversity and image quality. To generate the required text conditioning, we pass 4 of these selected views to a Vision-Language Model (VLM)~\cite{meta2025llama} to obtain a descriptive shape prompt. 

Finally, the 16 posed views, the filtered instance point cloud, and the VLM-generated text description are passed to ShapeR. Because the underlying flow-matching formulation is probabilistic and can yield variations in geometric fidelity, we run inference 4 times per object. This produces a set of four distinct metric candidate meshes per instance, ensuring we can subsequently select the highest-quality shape during the manual annotation phase.

\subsubsection{Annotation Process}

\begin{figure}[!]
\centering
\begin{tikzpicture}
\node(p0){\includegraphics[width=0.99\textwidth]{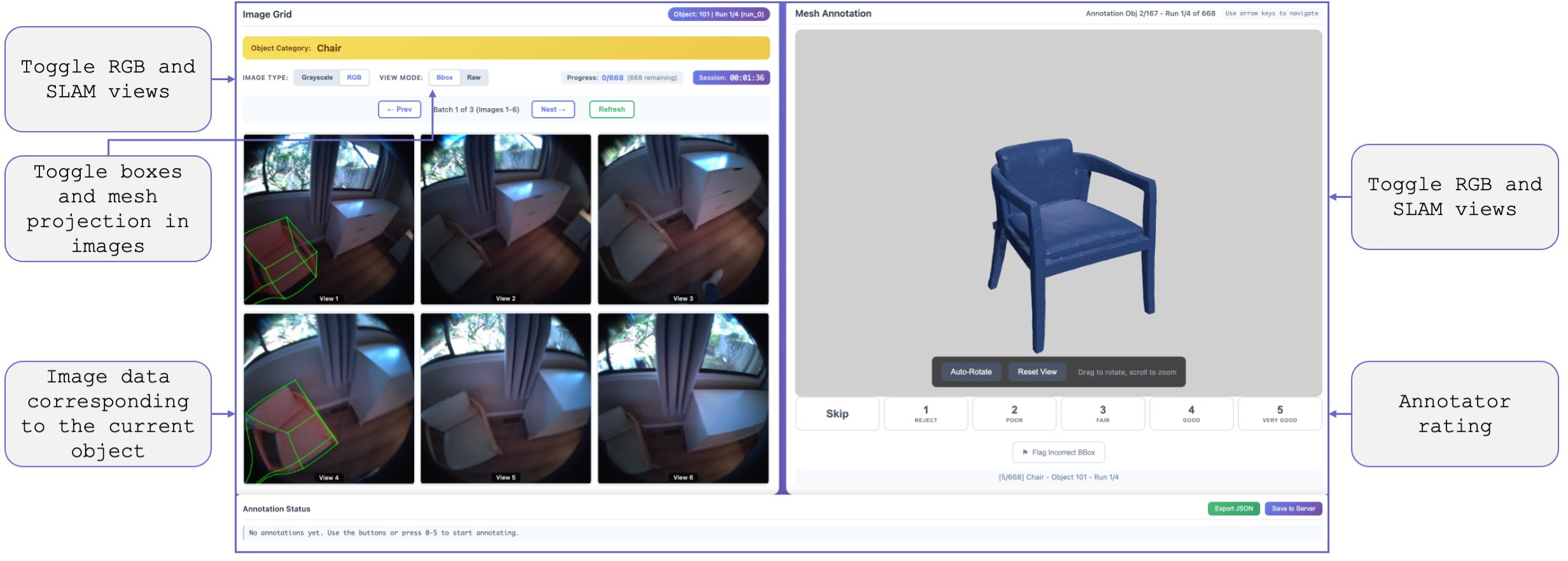}};
\end{tikzpicture}
\caption{\textbf{Shape Annotation Interface.} Our custom tool for verifying ShapeR-generated meshes. Annotators inspect an interactive 3D reconstruction (right) alongside multi-view reference images (left) with togglable 2D projections, rating the geometric fidelity on a 1 to 5 scale.}
\label{fig:shaper_annot_tool}
\end{figure}

Figure~\ref{fig:shaper_annot_tool} illustrates our custom annotation interface, designed to manually verify the fidelity of the ShapeR-generated 3D meshes. For a given bounding box, the interface displays the corresponding multi-view RGB images, grayscale images and relevant metadata such as the object category. To aid alignment verification, annotators can toggle 2D rendering of both the bounding box and the candidate mesh. An interactive 3D viewer on the right side of the screen allows users to inspect the generated mesh from arbitrary viewpoints. Using these visual references, annotators evaluate the reconstruction quality on a scale from 1 (poor) to 5 (excellent). This rating process is conducted for all four ShapeR candidate predictions per bounding box. We retain only the single highest-rated mesh for each instance, while strictly requiring a minimum quality score of 3 to be included in release. 

Our dataset-wide annotation pipeline is executed in three distinct phases. In the first phase, we exhaustively evaluate candidate meshes for the 47 basemap recordings, obtaining valid, high-quality shapes for 4547 out of 7171 total qualified object bounding boxes. In the second phase, similar to 3D bounding box propagation we propagate these validated geometries to the remaining sequence recordings. Here, a 3D shape is transferred from the basemap to a target recording if and only if the 3D IoU between the source and target bounding boxes is near perfect, ensuring strict spatial consistency. Finally, because this strict IoU threshold naturally disqualifies many potential transfers, the third phase addresses the remaining unpopulated bounding boxes. We run the full ShapeR inference and manual annotation protocol on these residual objects, retaining only the meshes that pass the established quality threshold.

%% file: sections/3.4-additional-data.tex
\subsection{Additional modalities for release}
\label{sect:moredata}

In \nymeriapp, we additionally release several data modalities that were not included in prior releases.
First, we release 47 new basemap recordings captured with Project Aria glasses, one per indoor venue.
These basemaps serve as the foundation for the 3D OBB and object-shape annotations described in~\cref{sect:nymeriapp}.
Second, we release all wristband videos.
Each wristband provides one RGB stream and two grayscale streams from forward-facing cameras.
In total, this corresponds to more than 600 hours of wrist-perspective video.
As with the Aria glasses footage, all wristband videos are anonymized.
Third, we release the full audio recordings from the Project Aria glasses, including both participants and observers, totaling more than 600 hours.
Finally, we backfill the original release with updated online calibration results, including RGB camera calibration.
We find that the improved online calibration yields more accurate downstream 3D reconstruction. 

%% file: sections/5-conclusion.tex
\section{Conclusion}
\label{sect:conclusion}

We introduced \nymeriapp, an enriched version of the \nymeria dataset that substantially expands its utility for egocentric human motion understanding and scene-aware reasoning.
\nymeriapp provides improved full-body motion annotations in widely used parametric formats (MHR and SMPL), large-scale 3D/2D oriented bounding box annotations for indoor objects and structural elements under both closed- and open-set taxonomies, and instance-level 3D object reconstructions.
In addition, we release previously withheld modalities---including basemap scans, wristband videos, and audio---together with updated calibration, enabling more comprehensive multimodal and geometric studies.

By combining accurate motion with dense scene context at scale, \nymeriapp bridges a key gap between in-the-wild egocentric perception, interaction modeling, and embodied AI.
We hope this release will support a broad range of future research, including environment-aware motion tracking and synthesis, open-world 3D object detection, fine-grained human--scene interaction analysis, and multimodal learning.

%% file: sections/app1-category.tex
\section{Fixed-Set Taxonomy for 3D OBB Annotations}
\label{app:fixed-taxonomy}

In this section, we describe the fixed-set taxonomy used to annotate commonly occurring objects (similar to the classes used in EFM3D~\cite{efm3d24}).

\textbf{Bed}. A bed is a surface intended for sleeping. This includes queen beds, bunk beds, bed frames without a mattress, mattresses without a bed frame, and cribs. A bunk bed is treated as a single object.

\textbf{Chair}. An object designed for a single person to sit on. Examples include beanbags, ottomans, three-legged chairs, and office chairs with caster wheels. Toilets are excluded because they serve a different function.

\textbf{Couch}. An object designed for one or more people to sit on, typically upholstered. Examples include L-shaped sectionals, hammocks, loveseats, and recliners. For sectional couches, annotators split the object into multiple sections to obtain tighter boxes.

\textbf{Door}. A static opening that a person can pass through and that can be closed by a door slab. Examples include closet doors, hallway entrances, shed entrances, storage-room doors, and bathroom entries. Very large doors designed for vehicles (e.g., garage doors) are excluded. This class does not include the door slab itself, which can articulate.

\textbf{Floor}. For simplicity and compatibility, we represent floors using 3D oriented bounding boxes, although planar polygons would be a more expressive representation.
Annotators align the top face of each floor OBB with the underlying floor plane.
For level, single-story venues, a single floor box typically covers the indoor space.
For multi-floor venues, we split the floor into multiple boxes; for outdoor areas with uneven ground, we approximate the surface using piecewise-planar boxes.
Annotators prioritize coverage of the ground beneath the Aria wearer so that downstream applications requiring height above ground can be supported throughout \nymeriapp.

\textbf{Lamp/Light}. A generic class that includes both standalone lights and light fixtures. For fixtures containing a light bulb, annotators provide the largest reasonable box; for example, a ceiling fan with lights is annotated as a single object covering the entire fan. Other examples include recessed lights, chandeliers, floor lamps, and soffit lights.

\textbf{Mirror}. An object with a reflective mirror surface, including full-body mirrors, vanity mirrors, personal face mirrors, and handheld mirrors. When a stand is present, annotators annotate only the mirror surface. Groups of nearby small mirrors may be annotated with a single box.

\textbf{Table}. An object whose primary purpose is to place items on (as opposed to storing items; see \textbf{Storage/Shelf}). Examples include dining tables, coffee tables, and desks. If an ottoman is used as a coffee table rather than a footrest, we annotate it as a table.

\textbf{Wall Art / Picture Frame}. A broad class covering decorative wall-mounted items and framed media, including wall art, posters, calendars, wall flags, and photo frames (whether wall-mounted or resting on furniture). Items such as light switches, wall shelves, vents, and temporarily hung personal objects are excluded. If multiple nearby pieces clearly form a single composite artwork, they may be annotated as one object.

\textbf{Window Opening}. Similar to \textbf{Door}, this class represents the static space occupied by a closed window, rather than the window itself, which may articulate. Examples include skylights and double-pane windows. Curtains are excluded when they occlude the window. The perimeter is defined as the space just inside the outermost window casing (when present). Windows embedded in doors are excluded because they are moveable.

\textbf{Plant}. This class includes potted plants (real or artificial), flowers in vases, shrubs, and trees. For indoor plants, we include the pot or vase. Outdoors, tightly clustered vegetation is difficult to separate, so we focus on more isolated instances. For trees, we annotate the full tree when observable; otherwise, we annotate only the trunk. Vines are excluded because they are poorly approximated by a 3D OBB.

\textbf{Storage/Shelf}. An object whose primary purpose is storage. Examples include closet rods, floating shelves, wall pegboards, window ledges, cabinets (e.g., under-sink or kitchen), shoe racks, dressers, mantles, shower nooks, built-in shelving, and toilet paper holders. For floating shelves, each shelf is annotated independently; for multi-shelf units, the entire unit is annotated as a single box. Temporary or frequently moved containers (e.g., laundry baskets, cardboard boxes, plastic tubs) are excluded.

\textbf{Screen/Display}. A digital display other than a handheld mobile phone. Examples include computer monitors, televisions, tablets, printer displays, arcade screens, and conference-room displays (including when powered off).

\textbf{Wall}. Similar to \textbf{Floor}, we approximate walls using 3D oriented bounding boxes for simplicity, although planar polygons would also be suitable.
We initialize each wall as a 12~cm thick volume; annotators may adjust thickness when warranted.
This class includes deck railings, pony walls, exterior siding, yard walls, fences, and angled staircase railings.
Curved walls are approximated using multiple smaller segments.
For walls with vaulted ceilings, we annotate a box that extends to the highest part of the wall.
Columns and pillars are excluded unless they are wider than 4~feet.
If an archway or cutout is part of the wall, we keep a single larger wall box that spans the opening rather than annotating sub-walls.

\textbf{Sink}. A fixed vessel for washing hands or objects, typically with a faucet (which is excluded from the annotation). Examples include bathroom vanity sinks, farmhouse kitchen sinks, and utility sinks. Exposed plumbing beneath the sink is excluded.

\textbf{Toilet}. We include the bowl and tank, but exclude the lid when it is open and extends above the tank.

\textbf{Refrigerator}. An appliance used to keep items cool, including mini fridges, wine fridges, and standalone freezers (e.g., in a garage).

\textbf{WasherDryer}. Standalone clothes washers, clothes dryers, and stackable washer--dryer towers. Dishwashers are excluded. If a door is open and extends beyond the main body, we exclude the door.

\textbf{Stairs}. We annotate each staircase as a sequence of steps. Single steps connecting subregions of a space are included (e.g., a step down to a garage). Non-walkable ledges (e.g., fireplace ledges) are excluded. For curved or angled staircases, we approximate the structure using a small number of segments.

%% file: sections/acknowledgement.tex
\section{Acknowledgement}
We thank Nan Yang, Jiaxi Jiang, Petr Kadlecek, and Jinlong Yang for dogfooding and support with human motion optimization. We also thank Dan Barnes and Raul Mur-Artal for early support on 3D bounding box annotations. We are grateful to Austin Kukay, Rowan Postyeni, Ruosha Pang, Mu Cheng, William Sun, Chen Zhang, Qinyue He, Aaron Deguzman, Yao Zhi, Luis Pesqueira, Abha Arora, and Rana Hayek for annotation support. Finally, we thank Fan Zhang, and Hauke Strasdat for general project support.

%% file: main.bib
@String(CVPR    = {IEEE/CVF Conference on Computer Vision and Pattern Recognition (CVPR)})

@String(ICCV    = {IEEE/CVF International Conference on Computer Vision (ICCV)})

@String(ECCV    = {European Conference on Computer Vision (ECCV)})

@String(BMVC    = {British Machine Vision Conference (BMVC)})

@String(ThreeDV = {International Conference on 3D Vision (3DV)})

@String(NeurIPS = {Advances in Neural Information Processing Systems (NeurIPS)})

@String(ICRA    = {IEEE International Conference on Robotics and Automation (ICRA)})

@String(SIGGA   = {ACM SIGGRAPH Asia Conference Proceedings})

@String(PAMI    = {IEEE Transactions on Pattern Analysis and Machine Intelligence})

@String(IJCV    = {International Journal of Computer Vision})

@String(TOG     = {ACM Transactions on Graphics})

@String(CGF     = {Computer Graphics Forum})

@misc{mhr25,
      title={{MHR}: Momentum Human Rig},
      author={Aaron Ferguson and Ahmed A. A. Osman and Berta Bescos and Carsten Stoll and Chris Twigg and Christoph Lassner and David Otte and Eric Vignola and Fabian Prada and Federica Bogo and Igor Santesteban and Javier Romero and Jenna Zarate and Jeongseok Lee and Jinhyung Park and Jinlong Yang and John Doublestein and Kishore Venkateshan and Kris Kitani and Ladislav Kavan and Marco Dal Farra and Matthew Hu and Matthew Cioffi and Michael Fabris and Michael Ranieri and Mohammad Modarres and Petr Kadlecek and Rawal Khirodkar and Rinat Abdrashitov and Romain Prévost and Roman Rajbhandari and Ronald Mallet and Russell Pearsall and Sandy Kao and Sanjeev Kumar and Scott Parrish and Shoou-I Yu and Shunsuke Saito and Takaaki Shiratori and Te-Li Wang and Tony Tung and Yichen Xu and Yuan Dong and Yuhua Chen and Yuanlu Xu and Yuting Ye and Zhongshi Jiang},
      year={2025},
      eprint={2511.15586},
      archivePrefix={arXiv},
      primaryClass={cs.GR},
      url={https://arxiv.org/abs/2511.15586},
}

@misc{momentum,
  key = {Momentum library},
  title = {\url{https://github.com/facebookresearch/momentum}},
}

@article{smpl15,
    author = {Loper, Matthew and Mahmood, Naureen and Romero, Javier and Pons-Moll, Gerard and Black, Michael J.},
    title = {{SMPL}: a skinned multi-person linear model},
    year = {2015},
    issue_date = {November 2015},
    publisher = {Association for Computing Machinery},
    address = {New York, NY, USA},
    volume = {34},
    number = {6},
    issn = {0730-0301},
    url = {https://doi.org/10.1145/2816795.2818013},
    doi = {10.1145/2816795.2818013},
    journal = TOG,
    month = nov,
    articleno = {248},
    numpages = {16}
}

@inproceedings{smplx19cvpr,
  title = {Expressive Body Capture: {3D} Hands, Face, and Body from a Single Image},
  author = {Pavlakos, Georgios and Choutas, Vasileios and Ghorbani, Nima and Bolkart, Timo and Osman, Ahmed A. A. and Tzionas, Dimitrios and Black, Michael J.},
  booktitle = CVPR,
  pages     = {10975--10985},
  year = {2019}
}

@inbook{smpl23,
    author = {Loper, Matthew and Mahmood, Naureen and Romero, Javier and Pons-Moll, Gerard and Black, Michael J.},
    title = {{SMPL}: A Skinned Multi-Person Linear Model},
    year = {2023},
    isbn = {9798400708978},
    publisher = {Association for Computing Machinery},
    address = {New York, NY, USA},
    edition = {1},
    url = {https://doi.org/10.1145/3596711.3596800},
    booktitle = {Seminal Graphics Papers: Pushing the Boundaries, Volume 2},
    articleno = {88},
    numpages = {16}
}

@article{efm3d24,
  title={EFM3D: A Benchmark for Measuring Progress Towards 3D Egocentric Foundation Models},
  author={Straub, Julian and DeTone, Daniel and Shen, Tianwei and Yang, Nan and Sweeney, Chris and Newcombe, Richard},
  journal={arXiv preprint arXiv:2406.10224},
  year={2024}
}

@misc{xsens,
  key = {Movella XSens},
  title = {\url{https://www.movella.com/motion-capture/xsens-link-specifications}},
}

@inproceedings{brazil2023omni3d,
  title={Omni3D: A Large Benchmark and Model for 3D Object Detection in the Wild},
  author={Brazil, Garrick and Kumar, Abhinav and Straub, Julian and Dzadzhu, Nikhila and Johnson, Justin and Ravi, Nikhila},
  booktitle=CVPR,
  year={2023}
}

@inproceedings{ahmadyan2021objectron,
  title={Objectron: A Large Scale Dataset of Object-Centric Videos in the Wild with Pose Annotations},
  author={Ahmadyan, Adel and Zhang, Liangkai and Ablavatski, Artsiom and Wei, Jianing and Grundmann, Matthias},
  booktitle=CVPR,
  year={2021}
}

@inproceedings{song2015sun,
  title={SUN RGB-D: A RGB-D Scene Understanding Benchmark Suite},
  author={Song, Shuran and Lichtenberg, Samuel P and Xiao, Jianxiong},
  booktitle=CVPR,
  year={2015}
}

@inproceedings{baruch2021arkitscenes,
  title={ARKitScenes: A Diverse Real-World Dataset for 3D Indoor Reconstruction},
  author={Baruch, Gilad and Chen, Zhuoyuan and Kostrikov, Ilya and Ma, Shuaibin and Yang, Zhuo and others},
  booktitle={NeurIPS Datasets and Benchmarks Track},
  year={2021}
}

@article{reizenstein2021common,
  title={Common Objects in 3D: Large-Scale Learning and Evaluation of Real-life 3D Category Reconstruction},
  author={Reizenstein, Jeremy and Shapovalov, Roman and Henzler, Philipp and Sbordone, Luca and Labatut, Patrick and Novotny, David},
  journal=ICCV,
  year={2021}
}

@inproceedings{ca1m24cvpr,
  title={Cubify anything: Scaling indoor 3d object detection},
  author={Lazarow, Justin and Griffiths, David and Kohavi, Gefen and Crespo, Francisco and Dehghan, Afshin},
  booktitle=CVPR,
  pages={22225--22233},
  year={2025}
}

@inproceedings{ego4d22cvpr,
    author    = {Grauman, Kristen and Westbury, Andrew and Byrne, Eugene and Chavis, Zachary and Furnari, Antonino and Girdhar, Rohit and Hamburger, Jackson and Jiang, Hao and Liu, Miao and Liu, Xingyu and Martin, Miguel and Nagarajan, Tushar and Radosavovic, Ilija and Ramakrishnan, Santhosh Kumar and Ryan, Fiona and Sharma, Jayant and Wray, Michael and Xu, Mengmeng and Xu, Eric Zhongcong and Zhao, Chen and Bansal, Siddhant and Batra, Dhruv and Cartillier, Vincent and Crane, Sean and Do, Tien and Doulaty, Morrie and Erapalli, Akshay and Feichtenhofer, Christoph and Fragomeni, Adriano and Fu, Qichen and Gebreselasie, Abrham and Gonz\'alez, Cristina and Hillis, James and Huang, Xuhua and Huang, Yifei and Jia, Wenqi and Khoo, Weslie and Kol\'a\v{r}, J\'achym and Kottur, Satwik and Kumar, Anurag and Landini, Federico and Li, Chao and Li, Yanghao and Li, Zhenqiang and Mangalam, Karttikeya and Modhugu, Raghava and Munro, Jonathan and Murrell, Tullie and Nishiyasu, Takumi and Price, Will and Ruiz, Paola and Ramazanova, Merey and Sari, Leda and Somasundaram, Kiran and Southerland, Audrey and Sugano, Yusuke and Tao, Ruijie and Vo, Minh and Wang, Yuchen and Wu, Xindi and Yagi, Takuma and Zhao, Ziwei and Zhu, Yunyi and Arbel\'aez, Pablo and Crandall, David and Damen, Dima and Farinella, Giovanni Maria and Fuegen, Christian and Ghanem, Bernard and Ithapu, Vamsi Krishna and Jawahar, C. V. and Joo, Hanbyul and Kitani, Kris and Li, Haizhou and Newcombe, Richard and Oliva, Aude and Park, Hyun Soo and Rehg, James M. and Sato, Yoichi and Shi, Jianbo and Shou, Mike Zheng and Torralba, Antonio and Torresani, Lorenzo and Yan, Mingfei and Malik, Jitendra},
    title     = {{Ego4D}: Around the World in 3,000 Hours of Egocentric Video},
    booktitle = CVPR,
    month     = {June},
    year      = {2022},
    pages     = {18995-19012}
}

@InProceedings{holoassist23iccv,
    author    = {Wang, Xin and Kwon, Taein and Rad, Mahdi and Pan, Bowen and Chakraborty, Ishani and Andrist, Sean and Bohus, Dan and Feniello, Ashley and Tekin, Bugra and Frujeri, Felipe Vieira and Joshi, Neel and Pollefeys, Marc},
    title     = {HoloAssist: an Egocentric Human Interaction Dataset for Interactive AI Assistants in the Real World},
    booktitle = ICCV,
    month     = {October},
    year      = {2023},
    pages     = {20270-20281}
}

@inproceedings{nymeria24eccv,
      title={Nymeria: A Massive Collection of Multimodal Egocentric Daily Motion in the Wild},
      author={Lingni Ma and Yuting Ye and Fangzhou Hong and Vladimir Guzov and Yifeng Jiang and Rowan Postyeni and Luis Pesqueira and Alexander Gamino and Vijay Baiyya and Hyo Jin Kim and Kevin Bailey and David Soriano Fosas and C. Karen Liu and Ziwei Liu and Jakob Engel and Renzo De Nardi and Richard Newcombe},
      booktitle= ECCV,
      year={2024},
      url={https://arxiv.org/abs/2406.09905},
}

@misc{egolife25,
      title={EgoLife: Towards Egocentric Life Assistant},
      author={Jingkang Yang and Shuai Liu and Hongming Guo and Yuhao Dong and Xiamengwei Zhang and Sicheng Zhang and Pengyun Wang and Zitang Zhou and Binzhu Xie and Ziyue Wang and Bei Ouyang and Zhengyu Lin and Marco Cominelli and Zhongang Cai and Yuanhan Zhang and Peiyuan Zhang and Fangzhou Hong and Joerg Widmer and Francesco Gringoli and Lei Yang and Bo Li and Ziwei Liu},
      year={2025},
      eprint={2503.03803},
      archivePrefix={arXiv},
      primaryClass={cs.CV},
      url={https://arxiv.org/abs/2503.03803},
}

@InProceedings{lamaria25iccv,
  author    = {Krishnan, Anusha and
               Liu, Shaohui and
               Sarlin, Paul-Edouard and
               Gentilhomme, Oscar and
               Caruso, David and
               Monge, Maurizio and
               Newcombe, Richard and
               Engel, Jakob and
               Pollefeys, Marc},
  title     = {Benchmarking Egocentric Visual-Inertial SLAM at City Scale},
  booktitle = ICCV,
  year      = {2025}
}

@inproceedings{lamar22eccv,
  author    = {Paul-Edouard Sarlin and
               Mihai Dusmanu and
               Johannes L. Schönberger and
               Pablo Speciale and
               Lukas Gruber and
               Viktor Larsson and
               Ondrej Miksik and
               Marc Pollefeys},
  title     = {{LaMAR: Benchmarking Localization and Mapping for Augmented Reality}},
  booktitle = ECCV,
  year      = {2022},
}

@article{egocom20pami,
  author={Curtis G. {Northcutt} and Shengxin {Zha} and Steven {Lovegrove} and Richard {Newcombe}},
  journal=PAMI,
  title={EgoCom: A Multi-person Multi-modal Egocentric Communications Dataset},
  year={2020},
  volume={},
  number={},
  pages={1-12},
  doi={10.1109/TPAMI.2020.3025105}
}

@inproceedings{gtea18eccv,
  title={In the eye of beholder: Joint learning of gaze and actions in first person video},
  author={Li, Yin and Liu, Miao and Rehg, James M},
  booktitle=ECCV,
  pages={619--635},
  year={2018}
}

@Article{trek22ijcv,
  author = {Dunnhofer, Matteo and Furnari, Antonino and Farinella, Giovanni Maria and Micheloni, Christian},
  title = {Visual Object Tracking in First Person Vision},
  journal = IJCV,
  year = {2022}
}

@inproceedings{egohands15iccv,
  title={Lending A Hand: Detecting Hands and Recognizing Activities in Complex Egocentric Interactions},
  author={Bambach, Sven and Lee, Stefan and Crandall, David and Yu, Chen},
  booktitle=ICCV,
  pages={196--204},
  year={2015}
}

@inproceedings{epickitchen18eccv,
   title={Scaling Egocentric Vision: The EPIC-KITCHENS Dataset},
   author={Damen, Dima and Doughty, Hazel and Farinella, Giovanni Maria  and Fidler, Sanja and Furnari, Antonino and Kazakos, Evangelos and Moltisanti, Davide and Munro, Jonathan and Perrett, Toby and Price, Will and Wray, Michael},
   booktitle=ECCV,
   year={2018}
}

@article{epickitchen20corr,
  title={Rescaling Egocentric Vision},
  author={Damen, Dima and Doughty, Hazel and Farinella, Giovanni Maria  and and Furnari, Antonino
          and Ma, Jian and Kazakos, Evangelos and Moltisanti, Davide and Munro, Jonathan
          and Perrett, Toby and Price, Will and Wray, Michael},
  journal   = {CoRR},
  volume    = {abs/2006.13256},
  year      = {2020},
  ee        = {http://arxiv.org/abs/2006.13256},
}

@inproceedings{h2o21cvpr,
  title={H2o: Two hands manipulating objects for first person interaction recognition},
  author={Kwon, Taein and Tekin, Bugra and St{\"u}hmer, Jan and Bogo, Federica and Pollefeys, Marc},
  booktitle={Proceedings of the IEEE/CVF international conference on computer vision},
  pages={10138--10148},
  year={2021}
}

@inproceedings{egoexo4d24cvpr,
    title={{Ego-Exo4D}: Understanding Skilled Human Activity from First- and Third-Person Perspectives},
    author={Kristen Grauman and Andrew Westbury and Lorenzo Torresani and Kris Kitani and Jitendra Malik and Triantafyllos Afouras and Kumar Ashutosh and Vijay Baiyya and Siddhant Bansal and Bikram Boote and Eugene Byrne and Zach Chavis and Joya Chen and Feng Cheng and Fu-Jen Chu and Sean Crane and Avijit Dasgupta and Jing Dong and Maria Escobar and Cristhian Forigua and Abrham Gebreselasie and Sanjay Haresh and Jing Huang and Md Mohaiminul Islam and Suyog Jain and Rawal Khirodkar and Devansh Kukreja and Kevin J Liang and Jia-Wei Liu and Sagnik Majumder and Yongsen Mao and Miguel Martin and Effrosyni Mavroudi and Tushar Nagarajan and Francesco Ragusa and Santhosh Kumar Ramakrishnan and Luigi Seminara and Arjun Somayazulu and Yale Song and Shan Su and Zihui Xue and Edward Zhang and Jinxu Zhang and Angela Castillo and Changan Chen and Xinzhu Fu and Ryosuke Furuta and Cristina Gonzalez and Prince Gupta and Jiabo Hu and Yifei Huang and Yiming Huang and Weslie Khoo and Anush Kumar and Robert Kuo and Sach Lakhavani and Miao Liu and Mi Luo and Zhengyi Luo and Brighid Meredith and Austin Miller and Oluwatumininu Oguntola and Xiaqing Pan and Penny Peng and Shraman Pramanick and Merey Ramazanova and Fiona Ryan and Wei Shan and Kiran Somasundaram and Chenan Song and Audrey Southerland and Masatoshi Tateno and Huiyu Wang and Yuchen Wang and Takuma Yagi and Mingfei Yan and Xitong Yang and Zecheng Yu and Shengxin Cindy Zha and Chen Zhao and Ziwei Zhao and Zhifan Zhu and Jeff Zhuo and Pablo Arbelaez and Gedas Bertasius and David Crandall and Dima Damen and Jakob Engel and Giovanni Maria Farinella and Antonino Furnari and Bernard Ghanem and Judy Hoffman and C. V. Jawahar and Richard Newcombe and Hyun Soo Park and James M. Rehg and Yoichi Sato and Manolis Savva and Jianbo Shi and Mike Zheng Shou and Michael Wray},
    year={2024},
    booktitle = CVPR,
}

@InProceedings{egopressure25cvpr,
    author    = {Zhao, Yiming and Kwon, Taein and Streli, Paul and Pollefeys, Marc and Holz, Christian},
    title     = {EgoPressure: A Dataset for Hand Pressure and Pose Estimation in Egocentric Vision},
    booktitle = CVPR,
    month     = {June},
    year      = {2025},
    pages     = {27727-27738}
}

@inproceedings{egobody22eccv,
   title = {{EgoBody}: Human Body Shape and Motion of Interacting People from Head-Mounted Devices},
   author = {Zhang, Siwei and Ma, Qianli and Zhang, Yan and Qian, Zhiyin and Kwon, Taein and Pollefeys, Marc and Bogo, Federica and Tang, Siyu},
   booktitle = ECCV,
   month = {Oct},
   year = {2022}
}

@inproceedings{hps21cvpr,
  title={Human poseitioning system (hps): 3d human pose estimation and self-localization in large scenes from body-mounted sensors},
  author={Guzov, Vladimir and Mir, Aymen and Sattler, Torsten and Pons-Moll, Gerard},
  booktitle=CVPR,
  pages={4318--4329},
  year={2021}
}

@inproceedings{egohuman23iccv,
    title={{EgoHumans}: An Egocentric 3D Multi-Human Benchmark},
    author={Rawal Khirodkar and Aayush Bansal and Lingni Ma and Richard Newcombe and Minh Vo and Kris Kitani},
    year={2023},
    booktitle = ICCV,
}

@inproceedings{harmony4d24neurips,
 author = {Khirodkar, Rawal and Song, Jyun-Ting and Cao, Jinkun and Luo, Zhengyi and Kitani, Kris},
 booktitle = NeurIPS,
 doi = {10.52202/079017-3407},
 editor = {A. Globerson and L. Mackey and D. Belgrave and A. Fan and U. Paquet and J. Tomczak and C. Zhang},
 pages = {107270--107285},
 publisher = {Curran Associates, Inc.},
 title = {Harmony4D: A Video Dataset for In-The-Wild Close Human Interactions},
 url = {https://proceedings.neurips.cc/paper_files/paper/2024/file/c20b843d0c6b1b40a8e6eb9a44e719c9-Paper-Datasets_and_Benchmarks_Track.pdf},
 volume = {37},
 year = {2024}
}

@conference{amass19iccv,
  title = {{AMASS}: Archive of Motion Capture as Surface Shapes},
  author = {Mahmood, Naureen and Ghorbani, Nima and Troje, Nikolaus F. and Pons-Moll, Gerard and Black, Michael J.},
  booktitle = ICCV,
  pages = {5442--5451},
  month = oct,
  year = {2019},
  month_numeric = {10}
}

@InProceedings{humanml3d22cvpr,
    author    = {Guo, Chuan and Zou, Shihao and Zuo, Xinxin and Wang, Sen and Ji, Wei and Li, Xingyu and Cheng, Li},
    title     = {Generating Diverse and Natural 3D Human Motions From Text},
    booktitle = CVPR,
    month     = {June},
    year      = {2022},
    pages     = {5152-5161}
}

@InProceedings{humoto25iccv,
      author    = {Lu, Jiaxin and Huang, Chun-Hao Paul and Bhattacharya, Uttaran and Huang, Qixing and Zhou, Yi},
      title     = {{HUMOTO}: A 4D Dataset of Mocap Human Object Interactions},
      booktitle = ICCV,
      month     = {October},
      year      = {2025},
      pages     = {10886-10897}
}

@inproceedings{behave22cvpr,
    title = {{BEHAVE}: Dataset and Method for Tracking Human Object Interactions},
    author={Bhatnagar, Bharat Lal and Xie, Xianghui and Petrov, Ilya and Sminchisescu, Cristian and Theobalt, Christian and Pons-Moll, Gerard},
    booktitle = CVPR,
    month = {jun},
    organization = {{IEEE}},
    year = {2022},
}

@inproceedings{bedlam23cvpr,
  title = {{BEDLAM}: A Synthetic Dataset of Bodies Exhibiting Detailed Lifelike Animated Motion},
  booktitle = CVPR,
  pages = {8726--8737},
  month = jun,
  year = {2023},
  author = {Black, Michael J. and Patel, Priyanka and Tesch, Joachim and Yang, Jinlong},
  doi = {10.1109/CVPR52729.2023.00843},
  url = {https://bedlam.is.tue.mpg.de/},
  month_numeric = {6}
}

@InProceedings{trumans24cvpr,
    author    = {Jiang, Nan and Zhang, Zhiyuan and Li, Hongjie and Ma, Xiaoxuan and Wang, Zan and Chen, Yixin and Liu, Tengyu and Zhu, Yixin and Huang, Siyuan},
    title     = {Scaling Up Dynamic Human-Scene Interaction Modeling},
    booktitle = CVPR,
    month     = {June},
    year      = {2024},
    pages     = {1737-1747}
}

@misc{parahome24,
    title={ParaHome: Parameterizing Everyday Home Activities Towards 3D Generative Modeling of Human-Object Interactions},
    author={Jeonghwan Kim and Jisoo Kim and Jeonghyeon Na and Hanbyul Joo},
    year={2024},
    eprint={2401.10232},
    archivePrefix={arXiv},
    primaryClass={cs.CV}
}

@misc{hdepic25,
      title={HD-EPIC: A Highly-Detailed Egocentric Video Dataset},
      author={Toby Perrett and Ahmad Darkhalil and Saptarshi Sinha and Omar Emara and Sam Pollard and Kranti Parida and Kaiting Liu and Prajwal Gatti and Siddhant Bansal and Kevin Flanagan and Jacob Chalk and Zhifan Zhu and Rhodri Guerrier and Fahd Abdelazim and Bin Zhu and Davide Moltisanti and Michael Wray and Hazel Doughty and Dima Damen},
      year={2025},
      eprint={2502.04144},
      archivePrefix={arXiv},
      primaryClass={cs.CV},
      url={https://arxiv.org/abs/2502.04144},
}

@inproceedings{adt23iccv,
  title={Aria digital twin: A new benchmark dataset for egocentric 3d machine perception},
  author={Pan, Xiaqing and Charron, Nicholas and Yang, Yongqian and Peters, Scott and Whelan, Thomas and Kong, Chen and Parkhi, Omkar and Newcombe, Richard and Ren, Yuheng Carl},
  booktitle=ICCV,
  pages={20133--20143},
  year={2023}
}

@misc{egovid5m24,
  title={{EgoVid-5M}: A Large-Scale Video-Action Dataset for Egocentric Video Generation},
  author={Xiaofeng Wang and Kang Zhao and Feng Liu and Jiayu Wang and Guosheng Zhao and Xiaoyi Bao and Zheng Zhu and Yingya Zhang and Xingang Wang},
  year={2024},
  eprint={2411.08380},
  archivePrefix={arXiv},
  primaryClass={cs.CV},
  url={https://arxiv.org/abs/2411.08380},
}

@misc{egodex25,
      title={EgoDex: Learning Dexterous Manipulation from Large-Scale Egocentric Video},
      author={Ryan Hoque and Peide Huang and David J. Yoon and Mouli Sivapurapu and Jian Zhang},
      year={2025},
      eprint={2505.11709},
      archivePrefix={arXiv},
      primaryClass={cs.CV},
      url={https://arxiv.org/abs/2505.11709},
}

@inproceedings{readingiwt25neurips,
  title={Reading recognition in the wild},
  author={Yang, Charig and Alam, Samiul and Siam, Shakhrul Iman and Proulx, Michael J and Mathias, Lambert and Somasundaram, Kiran and Pesqueira, Luis and Fort, James and Sheriffdeen, Sheroze and Parkhi, Omkar and others},
  booktitle=NeurIPS,
  year={2025}
}

@InProceedings{egotv23iccv,
    author    = {Hazra, Rishi and Chen, Brian and Rai, Akshara and Kamra, Nitin and Desai, Ruta},
    title     = {EgoTV: Egocentric Task Verification from Natural Language Task Descriptions},
    booktitle = ICCV,
    month     = {October},
    year      = {2023},
    pages     = {15417-15429}
}

@article{egoschema23neurips,
  title={Egoschema: A diagnostic benchmark for very long-form video language understanding},
  author={Mangalam, Karttikeya and Akshulakov, Raiymbek and Malik, Jitendra},
  journal=NeurIPS,
  volume={36},
  pages={46212--46244},
  year={2023}
}

@misc{wearvox25,
      title={WearVox: An Egocentric Multichannel Voice Assistant Benchmark for Wearables},
      author={Zhaojiang Lin and Yong Xu and Kai Sun and Jing Zheng and Yin Huang and Surya Teja Appini and Krish Narang and Renjie Tao and Ishan Kapil Jain and Siddhant Arora and Ruizhi Li and Yiteng Huang and Kaushik Patnaik and Wenfang Xu and Suwon Shon and Yue Liu and Ahmed A Aly and Anuj Kumar and Florian Metze and Xin Luna Dong},
      year={2025},
      eprint={2601.02391},
      archivePrefix={arXiv},
      primaryClass={cs.CL},
      url={https://arxiv.org/abs/2601.02391},
}

@inproceedings{motionmillion25iccv,
  title={Go to Zero: Towards Zero-shot Motion Generation with Million-scale Data},
  author={Ke Fan and Shunlin Lu and Minyue Dai and Runyi Yu and Lixing Xiao and Zhiyang Dou and Junting Dong and Lizhuang Ma and Jingbo Wang},
  year={2025},
  booktitle = ICCV,
}

@article{motionx23neurips,
  title={{Motion-X}: A large-scale 3d expressive whole-body human motion dataset},
  author={Lin, Jing and Zeng, Ailing and Lu, Shunlin and Cai, Yuanhao and Zhang, Ruimao and Wang, Haoqian and Zhang, Lei},
  journal=NeurIPS,
  volume={36},
  pages={25268--25280},
  year={2023}
}

@article{motionxpp25,
  title={{Motion-X++}: A Large-Scale Multimodal 3D Whole-body Human Motion Dataset},
  author={Zhang, Yuhong and Lin, Jing and Zeng, Ailing and Wu, Guanlin and Lu, Shunlin and Fu, Yurong and Cai, Yuanhao and Zhang, Ruimao and Wang, Haoqian and Zhang, Lei},
  journal={arXiv preprint arXiv:2501.05098},
  year={2025}
}

@article{smplerx23neurips,
  title={Smpler-x: Scaling up expressive human pose and shape estimation},
  author={Cai, Zhongang and Yin, Wanqi and Zeng, Ailing and Wei, Chen and Sun, Qingping and Yanjun, Wang and Pang, Hui En and Mei, Haiyi and Zhang, Mingyuan and Zhang, Lei and others},
  journal=NeurIPS,
  volume={36},
  pages={11454--11468},
  year={2023}
}

@inproceedings{slahmr23cvpr,
  title={Decoupling human and camera motion from videos in the wild},
  author={Ye, Vickie and Pavlakos, Georgios and Malik, Jitendra and Kanazawa, Angjoo},
  booktitle=CVPR,
  pages={21222--21232},
  year={2023}
}

@article{prompthmr25cvpr,
  title={PromptHMR: Promptable Human Mesh Recovery},
  author={Wang, Yufu and Sun, Yu and Patel, Priyanka and Daniilidis, Kostas and Black, Michael J and Kocabas, Muhammed},
  journal=CVPR,
  year={2025}
}

@inproceedings{multihmr24eccv,
    title={Multi-HMR: Multi-Person Whole-Body Human Mesh Recovery in a Single Shot},
    author={Baradel*, Fabien and
            Armando, Matthieu and
            Galaaoui, Salma and
            Br{\'e}gier, Romain and
            Weinzaepfel, Philippe and
            Rogez, Gr{\'e}gory and
            Lucas*, Thomas
            },
    booktitle=ECCV,
    year={2024}
}

@InProceedings{satmhr25cvpr,
    author    = {Su, Chi and Ma, Xiaoxuan and Su, Jiajun and Wang, Yizhou},
    title     = {SAT-HMR: Real-Time Multi-Person 3D Mesh Estimation via Scale-Adaptive Tokens},
    booktitle = CVPR,
    month     = {June},
    year      = {2025},
    pages     = {16796-16806}
}

@inproceedings{frankmocap21iccv,
  title={Frankmocap: A monocular 3d whole-body pose estimation system via regression and integration},
  author={Rong, Yu and Shiratori, Takaaki and Joo, Hanbyul},
  booktitle=ICCV,
  pages={1749--1759},
  year={2021}
}

@article{physcap20tog,
  title={Physcap: Physically plausible monocular 3d motion capture in real time},
  author={Shimada, Soshi and Golyanik, Vladislav and Xu, Weipeng and Theobalt, Christian},
  journal=TOG,
  volume={39},
  number={6},
  pages={1--16},
  year={2020},
  publisher={ACM New York, NY, USA}
}

@article{divatrack24cgf,
  author = {Yang, Dongseok and Kang, Jiho and Ma, Lingni and Greer, Joseph and Ye, Vickie and Lee, Sung-Hee},
  title = {DivaTrack: Diverse Bodies and Motions from Acceleration-Enhanced 3-Point Trackers},
  journal = CGF,
  volume = {43},
  number = {2},
  pages = {e15057},
  year = {2024},
  doi = {https://doi.org},
}

@inproceedings{empose21iccv,
  title={Em-pose: 3d human pose estimation from sparse electromagnetic trackers},
  author={Kaufmann, Manuel and Zhao, Yi and Tang, Chengcheng and Tao, Lingling and Twigg, Christopher and Song, Jie and Wang, Robert and Hilliges, Otmar},
  booktitle=ICCV,
  pages={11510--11520},
  year={2021}
}

@inproceedings{emdb23iccv,
  title={Emdb: The electromagnetic database of global 3d human pose and shape in the wild},
  author={Kaufmann, Manuel and Song, Jie and Guo, Chen and Shen, Kaiyue and Jiang, Tianjian and Tang, Chengcheng and Z{\'a}rate, Juan Jos{\'e} and Hilliges, Otmar},
  booktitle=ICCV,
  pages={14632--14643},
  year={2023}
}

@inproceedings{mocapee24cvpr,
  title={Mocap everyone everywhere: Lightweight motion capture with smartwatches and a head-mounted camera},
  author={Lee, Jiye and Joo, Hanbyul},
  booktitle=CVPR,
  pages={1091--1100},
  year={2024}
}

@inproceedings{3dpw18eccv,
  title={Recovering accurate 3d human pose in the wild using imus and a moving camera},
  author={Von Marcard, Timo and Henschel, Roberto and Black, Michael J and Rosenhahn, Bodo and Pons-Moll, Gerard},
  booktitle=ECCV,
  pages={601--617},
  year={2018}
}

@InProceedings{gorp25cvpr,
    author    = {Barquero, German and Bertsch, Nadine and Marramreddy, Manojkumar and Chac\'on, Carlos and Arcadu, Filippo and Rigual, Ferran and He, Nicky Sijia and Palmero, Cristina and Escalera, Sergio and Ye, Yuting and Kips, Robin},
    title     = {From Sparse Signal to Smooth Motion: Real-Time Motion Generation with Rolling Prediction Models},
    booktitle = CVPR,
    month     = {June},
    year      = {2025},
    pages     = {1850-1860}
}

@inproceedings{egomimic25icra,
  title={Egomimic: Scaling imitation learning via egocentric video},
  author={Kareer, Simar and Patel, Dhruv and Punamiya, Ryan and Mathur, Pranay and Cheng, Shuo and Wang, Chen and Hoffman, Judy and Xu, Danfei},
  booktitle=ICRA,
  pages={13226--13233},
  year={2025},
  organization={IEEE}
}

@inproceedings{panoptic15iccv,
  title={Panoptic studio: A massively multiview system for social motion capture},
  author={Joo, Hanbyul and Liu, Hao and Tan, Lei and Gui, Lin and Nabbe, Bart and Matthews, Iain and Kanade, Takeo and Nobuhara, Shohei and Sheikh, Yaser},
  booktitle=ICCV,
  pages={3334--3342},
  year={2015}
}

@inproceedings{egogen24cvpr,
  title={Egogen: An egocentric synthetic data generator},
  author={Li, Gen and Zhao, Kaifeng and Zhang, Siwei and Lyu, Xiaozhong and Dusmanu, Mihai and Zhang, Yan and Pollefeys, Marc and Tang, Siyu},
  booktitle=CVPR,
  pages={14497--14509},
  year={2024}
}

@misc{embody3d25,
      title={Embody 3D: A Large-scale Multimodal Motion and Behavior Dataset},
      author={Claire McLean and Makenzie Meendering and Tristan Swartz and Orri Gabbay and Alexandra Olsen and Rachel Jacobs and Nicholas Rosen and Philippe de Bree and Tony Garcia and Gadsden Merrill and Jake Sandakly and Julia Buffalini and Neham Jain and Steven Krenn and Moneish Kumar and Dejan Markovic and Evonne Ng and Fabian Prada and Andrew Saba and Siwei Zhang and Vasu Agrawal and Tim Godisart and Alexander Richard and Michael Zollhoefer},
      year={2025},
      eprint={2510.16258},
      archivePrefix={arXiv},
      primaryClass={cs.CV},
      url={https://arxiv.org/abs/2510.16258},
}

@article{circle23cvpr,
    title={{CIRCLE}: Capture In Rich Contextual Environments},
    author={Joao Pedro Araujo and Jiaman Li and Karthik Vetrivel and Rishi Agarwal and Deepak Gopinath and Jiajun Wu and Alexander Clegg and C. Karen Liu},
    year={2023},
    journal=CVPR,
}

@inproceedings{unrealego22eccv,
	title = {{UnrealEgo}: A New Dataset for Robust Egocentric 3D Human Motion Capture},
	author = {Akada, Hiroyasu and Wang, Jian and Shimada, Soshi and Takahashi, Masaki and Theobalt, Christian and Golyanik, Vladislav},
	booktitle = ECCV,
	year = {2022}
}

@article{openego25,
  title={Openego: A large-scale multimodal egocentric dataset for dexterous manipulation},
  author={Jawaid, Ahad and Xiang, Yu},
  journal={arXiv preprint arXiv:2509.05513},
  year={2025}
}

@misc{unihand226,
  title={Being-H0.5: Scaling Human-Centric Robot Learning for Cross-Embodiment Generalization},
  author={Hao Luo and Ye Wang and Wanpeng Zhang and Sipeng Zheng and Ziheng Xi and Chaoyi Xu and Haiweng Xu and Haoqi Yuan and Chi Zhang and Yiqing Wang and Yicheng Feng and Zongqing Lu},
  year={2026},
  eprint={2601.12993},
  archivePrefix={arXiv},
  primaryClass={cs.RO},
  url={https://arxiv.org/abs/2601.12993},
}

@inproceedings{scannet17cvpr,
  title={ScanNet: Richly-annotated 3D Reconstructions of Indoor Scenes},
  author={Dai, Angela and Chang, Angel X. and Savva, Manolis and Halber, Maciej and Funkhouser, Thomas and Nie{\ss}ner, Matthias},
  booktitle=CVPR,
  year={2017}
}

@inproceedings{scannet++23cvpr,
  title={Scannet++: A high-fidelity dataset of 3d indoor scenes},
  author={Yeshwanth, Chandan and Liu, Yueh-Cheng and Nie{\ss}ner, Matthias and Dai, Angela},
  booktitle=ICCV,
  pages={12--22},
  year={2023}
}

@article{replica19arxiv,
  title =   {The {R}eplica Dataset: A Digital Replica of Indoor Spaces},
  author =  {Julian Straub and Thomas Whelan and Lingni Ma and Yufan Chen and Erik Wijmans and Simon Green and Jakob J. Engel and Raul Mur-Artal and Carl Ren and Shobhit Verma and Anton Clarkson and Mingfei Yan and Brian Budge and Yajie Yan and Xiaqing Pan and June Yon and Yuyang Zou and Kimberly Leon and Nigel Carter and Jesus Briales and  Tyler Gillingham and  Elias Mueggler and Luis Pesqueira and Manolis Savva and Dhruv Batra and Hauke M. Strasdat and Renzo De Nardi and Michael Goesele and Steven Lovegrove and Richard Newcombe },
  journal = {arXiv preprint arXiv:1906.05797},
  year =    {2019}
}

@inproceedings{ase24eccv,
  title={Scenescript: Reconstructing scenes with an autoregressive structured language model},
  author={Avetisyan, Armen and Xie, Christopher and Howard-Jenkins, Henry and Yang, Tsun-Yi and Aroudj, Samir and Patra, Suvam and Zhang, Fuyang and Frost, Duncan and Holland, Luke and Orme, Campbell and others},
  booktitle=ECCV,
  pages={247--263},
  year={2024},
  organization={Springer}
}

@inproceedings{mse25,
  author = {Klinghoffer, Tzofi and
  Somasundaram, Siddharth and
  Xiang, Xiaoyu and
  Fan, Yuchen and
  Richardt, Christian and
  Dave, Akshat and
  Raskar, Ramesh and
  Ranjan, Rakesh},
  title = {{Shoot-Bounce-3D}: Single-Shot Occlusion-Aware
  {3D} from Lidar by Decomposing Two-Bounce Light},
  booktitle = SIGGA,
  year = {2025},
  url = {https://shoot-bounce-3d.github.io},
}

@inproceedings{objaverse23cvpr,
  title={Objaverse: A universe of annotated 3d objects},
  author={Deitke, Matt and Schwenk, Dustin and Salvador, Jordi and Weihs, Luca and Michel, Oscar and VanderBilt, Eli and Schmidt, Ludwig and Ehsani, Kiana and Kembhavi, Aniruddha and Farhadi, Ali},
  booktitle=CVPR,
  pages={13142--13153},
  year={2023}
}

@inproceedings{objaversexl23neurips,
  author = {Deitke, Matt and Liu, Ruoshi and Wallingford, Matthew and Ngo, Huong and Michel, Oscar and Kusupati, Aditya and Fan, Alan and Laforte, Christian and Voleti, Vikram and Gadre, Samir Yitzhak and VanderBilt, Eli and Kembhavi, Aniruddha and Vondrick, Carl and Gkioxari, Georgia and Ehsani, Kiana and Schmidt, Ludwig and Farhadi, Ali},
  title = {Objaverse-XL: a universe of 10M+ 3D objects},
  year = {2023},
  publisher = {Curran Associates Inc.},
  address = {Red Hook, NY, USA},
  booktitle = NeurIPS,
  articleno = {1554},
  numpages = {15},
  location = {New Orleans, LA, USA},
  series = {NIPS '23}
}

@inproceedings{egoobject23iccv,
  title={EgoObjects: A Large-Scale Egocentric Dataset for Fine-Grained Object Understanding},
  author={Zhu, Chenchen and Xiao, Fanyi and Alvarado, Andrés and Babaei, Yasmine and Hu, Jiabo and El-Mohri, Hichem and Chang, Sean and Sumbaly, Roshan and Yan, Zhicheng},
  booktitle=ICCV,
  year={2023}
}

@inproceedings{dtc25cvpr,
    author    = {Dong, Zhao and Chen, Ka and Lv, Zhaoyang and Yu, Hong-Xing and Zhang, Yunzhi and Zhang, Cheng and Zhu, Yufeng and Tian, Stephen and Li, Zhengqin and Moffatt, Geordie and Christofferson, Sean and Fort, James and Pan, Xiaqing and Yan, Mingfei and Wu, Jiajun and Ren, Carl Yuheng and Newcombe, Richard},
    title     = {Digital Twin Catalog: A Large-Scale Photorealistic 3D Object Digital Twin Dataset},
    booktitle = CVPR,
    month     = {June},
    year      = {2025},
    pages     = {753-763}
}

@article{hot3d25cvpr,
  title={{HOT3D}: Hand and Object Tracking in {3D} from Egocentric Multi-View Videos},
  author={Banerjee, Prithviraj and Shkodrani, Sindi and Moulon, Pierre and Hampali, Shreyas and Han, Shangchen and Zhang, Fan and Zhang, Linguang and Fountain, Jade and Miller, Edward and Basol, Selen and Newcombe, Richard and Wang, Robert and Engel, Jakob Julian and Hodan, Tomas},
  journal=CVPR,
  year={2025}
}

@inproceedings{assembly10122cvpr,
  title={Assembly101: A large-scale multi-view video dataset for understanding procedural activities},
  author={Sener, Fadime and Chatterjee, Dibyadip and Shelepov, Daniel and He, Kun and Singhania, Dipika and Wang, Robert and Yao, Angela},
  booktitle=CVPR,
  pages={21096--21106},
  year={2022}
}

@inproceedings{mmcsg24,
  title={The CHiME-8 MMCSG Challenge: Multi-modal conversations in smart glasses},
  author={Zmolikova, Katerina and Merello, Simone and Kalgaonkar, Kaustubh and Lin, Ju and Moritz, Niko and Ma, Pingchuan and Sun, Ming and Chen, Honglie and Saliou, Antoine and Petridis, Stavros and others},
  booktitle={8th International Workshop on Speech Processing in Everyday Environments (CHiME)},
  pages={7--12},
  year={2024}
}

@inproceedings{dexycb21cvpr,
  author    = {Yu-Wei Chao and Wei Yang and Yu Xiang and Pavlo Molchanov and Ankur Handa and Jonathan Tremblay and Yashraj S. Narang and Karl {Van Wyk} and Umar Iqbal and Stan Birchfield and Jan Kautz and Dieter Fox},
  booktitle = CVPR,
  title     = {{DexYCB}: A Benchmark for Capturing Hand Grasping of Objects},
  year      = {2021},
}

@inproceedings{grab20eccv,
  title = {{GRAB}: A Dataset of Whole-Body Human Grasping of Objects},
  author = {Taheri, Omid and Ghorbani, Nima and Black, Michael J. and Tzionas, Dimitrios},
  booktitle = ECCV,
  year = {2020},
  url = {https://grab.is.tue.mpg.de}
}

@inproceedings{gigahands25cvpr,
  title={Gigahands: A massive annotated dataset of bimanual hand activities},
  author={Fu, Rao and Zhang, Dingxi and Jiang, Alex and Fu, Wanjia and Funk, Austin and Ritchie, Daniel and Sridhar, Srinath},
  booktitle=CVPR,
  pages={17461--17474},
  year={2025}
}

@misc{aria23surreal,
      title={{Project Aria}: A New Tool for Egocentric Multi-Modal {AI} Research},
      author={Jakob Engel and Kiran Somasundaram and Michael Goesele and Albert Sun and Alexander Gamino and Andrew Turner and Arjang Talattof and Arnie Yuan and Bilal Souti and Brighid Meredith and Cheng Peng and Chris Sweeney and Cole Wilson and Dan Barnes and Daniel DeTone and David Caruso and Derek Valleroy and Dinesh Ginjupalli and Duncan Frost and Edward Miller and Elias Mueggler and Evgeniy Oleinik and Fan Zhang and Guruprasad Somasundaram and Gustavo Solaira and Harry Lanaras and Henry Howard-Jenkins and Huixuan Tang and Hyo Jin Kim and Jaime Rivera and Ji Luo and Jing Dong and Julian Straub and Kevin Bailey and Kevin Eckenhoff and Lingni Ma and Luis Pesqueira and Mark Schwesinger and Maurizio Monge and Nan Yang and Nick Charron and Nikhil Raina and Omkar Parkhi and Peter Borschowa and Pierre Moulon and Prince Gupta and Raul Mur-Artal and Robbie Pennington and Sachin Kulkarni and Sagar Miglani and Santosh Gondi and Saransh Solanki and Sean Diener and Shangyi Cheng and Simon Green and Steve Saarinen and Suvam Patra and Tassos Mourikis and Thomas Whelan and Tripti Singh and Vasileios Balntas and Vijay Baiyya and Wilson Dreewes and Xiaqing Pan and Yang Lou and Yipu Zhao and Yusuf Mansour and Yuyang Zou and Zhaoyang Lv and Zijian Wang and Mingfei Yan and Carl Ren and Renzo De Nardi and Richard Newcombe},
      year={2023},
      eprint={2308.13561},
      archivePrefix={arXiv},
      primaryClass={cs.HC}
}

@misc{hololens,
    title = {Microsoft {HoloLens}},
    url = {https://learn.microsoft.com/en-us/hololens/},
}

@misc{rayban,
    title = {{Ray-Ban Meta} Smart Glasses},
    url = {https://www.meta.com/smart-glasses/},
}

@misc{vuzix,
    title={Vuzix Smart Glasses},
    url={https://www.vuzix.com/pages/smart-glasses},
}

@misc{quest,
    title={{Meta Quest}},
    url={https://www.meta.com/quest/},
}

@misc{applevisionpro,
    title={Apple {Vision Pro}},
    url={https://www.apple.com/apple-vision-pro/},
}

@misc{vive,
    title = {{HTC VIVE}},
    url = {vive.com}
}

@misc{pytorch,
    title = {{Pytorch library}},
    url = {https://pytorch.org/}
}

@misc{accad,
  title           = {{ACCAD MoCap Dataset}},
  author          = {{Advanced Computing Center for the Arts and Design}},
  url             = {https://accad.osu.edu/research/motion-lab/mocap-system-and-data}
}

@inproceedings{bmlhandball15,
  author          = {Helm, Fabian and Troje, Nikolaus and Reiser, Mathias and Munzert, Jörn},
  year            = {2015},
  month           = {01},
  pages           = {},
  title           = {Bewegungsanalyse getäuschter und nicht-getäuschter 7m-Würfe im Handball},
  journal         = {47. Jahrestagung der Arbeitsgemeinschaft für Sportpsychologie, Freiburg.}
}

@article{bmlmovi20,
  title           = {{MoVi}: A Large Multipurpose Motion and Video Dataset},
  author          = {Saeed Ghorbani and Kimia Mahdaviani and Anne Thaler and Konrad Kording and Douglas James Cook and Gunnar Blohm and Nikolaus F. Troje},
  year            = {2020},
  journal         = {arXiv preprint arXiv: 2003.01888}
}

@article{bmlrub02,
  title           = {Decomposing Biological Motion: {A} Framework for Analysis and Synthesis of Human Gait Patterns},
  author          = {Troje, Nikolaus F.},
  year            = 2002,
  month           = sep,
  journal         = {Journal of Vision},
  volume          = 2,
  number          = 5,
  pages           = {2--2},
  doi             = {10.1167/2.5.2},
  month_numeric   = 9
}

@misc{cmumocap,
  title           = {{CMU MoCap Dataset}},
  author          = {{Carnegie Mellon University}},
  url             = {http://mocap.cs.cmu.edu}
}

@article{dancedb,
  author          = {Aristidou, Andreas and Shamir, Ariel and Chrysanthou, Yiorgos},
  title           = {Digital Dance Ethnography: {O}rganizing Large Dance Collections},
  journal         = {J. Comput. Cult. Herit.},
  issue_date      = {January 2020},
  volume          = {12},
  number          = {4},
  month           = nov,
  year            = {2019},
  issn            = {1556-4673},
  articleno       = {29},
  numpages        = {27},
  url             = {https://doi.org/10.1145/3344383},
  doi             = {10.1145/3344383},
  acmid           = {},
  publisher       = {Association for Computing Machinery},
  address         = {New York, NY, USA},
}

@inproceedings{dfaust,
  title           = {Dynamic {FAUST}: {R}egistering Human Bodies in Motion},
  author          = {Bogo, Federica and Romero, Javier and Pons-Moll, Gerard and Black, Michael J.},
  booktitle       = {IEEE Conf. on Computer Vision and Pattern Recognition (CVPR)},
  month           = jul,
  year            = {2017},
  month_numeric   = {7}
}

@misc{eyesjapandataset,
  title           = {{Eyes Japan MoCap Dataset}},
  author          = {Eyes JAPAN Co. Ltd.},
  url             = {http://mocapdata.com}
}

@inproceedings{contactdb19cvpr,
  title           = {{ContactDB}: Analyzing and Predicting Grasp Contact via Thermal Imaging},
  author          = {Brahmbhatt, Samarth and Ham, Cusuh and Kemp, Charles C. and Hays, James},
  booktitle       = CVPR,
  year            = {2019},
  url             = {https://contactdb.cc.gatech.edu}
}

@techreport{hdm05,
  author          = {M. M\"{u}ller and T. R\"{o}der and M. Clausen and B. Eberhardt and B. Kr\"{u}ger and A. Weber},
  title           = {Documentation Mocap Database HDM05},
  number          = {CG-2007-2},
  year            = {2007},
  month           = {June},
  institution     = {Universit\"{a}t Bonn},
  issn            = {1610-8892}
}

@article{human4d20,
  title           = {HUMAN4D: A Human-Centric Multimodal Dataset for Motions and Immersive Media},
  author          = {Chatzitofis, Anargyros and Saroglou, Leonidas and Boutis, Prodromos and Drakoulis, Petros and Zioulis, Nikolaos and Subramanyam, Shishir and Kevelham, Bart and Charbonnier, Caecilia and Cesar, Pablo and Zarpalas, Dimitrios and others},
  journal         = {IEEE Access},
  volume          = {8},
  pages           = {176241--176262},
  year            = {2020},
  publisher       = {IEEE}
}

@article{humaneva10ijcv,
  title           = {{HumanEva}: Synchronized video and motion capture dataset and baseline algorithm for evaluation of articulated human motion},
  author          = {Sigal, L. and Balan, A. and Black, M. J.},
  journal         = IJCV,
  volume          = {87},
  number          = {1},
  pages           = {4--27},
  publisher       = {Springer Netherlands},
  month           = mar,
  year            = {2010},
  doi             = {},
  month_numeric   = {3}
}

@inproceedings{kit15icra,
  author          = {Christian Mandery and \"Omer Terlemez and Martin Do and Nikolaus Vahrenkamp and Tamim Asfour},
  title           = {The {KIT} Whole-Body Human Motion Database},
  booktitle       = ICRA,
  pages           = {329--336},
  year            = {2015},
}

@article{kit16,
  author          = {Christian Mandery and \"Omer Terlemez and Martin Do and Nikolaus Vahrenkamp and Tamim Asfour},
  title           = {Unifying Representations and Large-Scale Whole-Body Motion Databases for Studying Human Motion},
  pages           = {796--809},
  volume          = {32},
  number          = {4},
  journal         = {IEEE Transactions on Robotics},
  year            = {2016},
}

@inproceedings{kit21,
  author          = {Franziska Krebs and Andre Meixner and Isabel Patzer and Tamim Asfour},
  title           = {The {KIT} Bimanual Manipulation Dataset},
  booktitle       = {IEEE/RAS International Conference on Humanoid Robots (Humanoids)},
  pages           = {499--506},
  year            = {2021},
}

@inproceedings{moyo23cvpr,
  title           = {{3D} Human Pose Estimation via Intuitive Physics},
  author          = {Tripathi, Shashank and M{\"u}ller, Lea and Huang, Chun-Hao P. and Taheri Omid and Black, Michael J. and Tzionas, Dimitrios},
  booktitle       = CVPR,
  month           = {June},
  year            = {2023}
}

@article{mosh14tog,
  title           = {{MoSh}: Motion and Shape Capture from Sparse Markers},
  author          = {Loper, Matthew M. and Mahmood, Naureen and Black, Michael J.},
  address         = {New York, NY, USA},
  publisher       = {ACM},
  month           = nov,
  number          = {6},
  volume          = {33},
  pages           = {220:1--220:13},
  journal         = {ACM Transactions on Graphics, (Proc. SIGGRAPH Asia)},
  url             = {http://doi.acm.org/10.1145/2661229.2661273},
  year            = {2014},
  doi             = {10.1145/2661229.2661273}
}

@inproceedings{poseprior15cvpr,
  title           = {Pose-Conditioned Joint Angle Limits for {3D} Human Pose Reconstruction},
  author          = {Akhter, Ijaz and Black, Michael J.},
  booktitle       = CVPR,
  month           = jun,
  year            = {2015}
}

@misc{sfu,
  title           = {{SFU Motion Capture Database}},
  author          = {Simon Fraser University and National University of Singapore},
  url             = {http://mocap.cs.sfu.ca/}
}

@inproceedings{soma21iccv,
  title           = {{SOMA}: Solving Optical Marker-Based MoCap Automatically},
  author          = {Ghorbani, Nima and Black, Michael J.},
  booktitle       = {Proc. International Conference on Computer Vision (ICCV)},
  pages           = {11117--11126},
  month           = oct,
  year            = {2021},
  doi             = {},
  month_numeric   = {10}
}

@inproceedings{tcdhand12,
  author          = {Ludovic Hoyet and Kenneth Ryall and Rachel McDonnell and Carol O'Sullivan},
  title           = {Sleight of Hand: Perception of Finger Motion from Reduced Marker Sets},
  booktitle       = {Proceedings of the ACM SIGGRAPH Symposium on Interactive 3D Graphics and Games},
  year            = {2012},
  pages           = {79--86},
  doi             = {10.1145/2159616.2159629}
}

@inproceedings{totalcapture17bmvc,
  author          = {Trumble, Matt and Gilbert, Andrew and Malleson, Charles and  Hilton, Adrian and Collomosse, John},
  title           = {{Total Capture}: 3D Human Pose Estimation Fusing Video and Inertial Sensors},
  booktitle       = {2017 British Machine Vision Conference (BMVC)},
  year            = {2017}
}

@inproceedings{wheelposer24,
  title={WheelPoser: Sparse-IMU Based Body Pose Estimation for Wheelchair Users},
  author={Li, Yunzhi and Mollyn, Vimal and Yuan, Kuang and Carrington, Patrick},
  booktitle={Proceedings of the 26th International ACM SIGACCESS Conference on Computers and Accessibility},
  pages={1--17},
  year={2024}
}

@article{handeye17,
  title={Least-squares rigid motion using svd},
  author={Sorkine-Hornung, Olga and Rabinovich, Michael},
  journal={Computing},
  volume={1},
  number={1},
  pages={1--5},
  year={2017}
}

@inproceedings{hmd2253dv,
  title={{HMD$^2$}: Environment-aware motion generation from single egocentric head-mounted device},
  author={Guzov, Vladimir and Jiang, Yifeng and Hong, Fangzhou and Pons-Moll, Gerard and Newcombe, Richard and Liu, C Karen and Ye, Yuting and Ma, Lingni},
  booktitle=ThreeDV,
  pages={1394--1405},
  year={2025},
  organization={IEEE}
}

@InProceedings{egolm25cvpr,
    author    = {Hong, Fangzhou and Guzov, Vladimir and Kim, Hyo Jin and Ye, Yuting and Newcombe, Richard and Liu, Ziwei and Ma, Lingni},
    title     = {EgoLM: Multi-Modal Language Model of Egocentric Motions},
    booktitle = CVPR,
    month     = {June},
    year      = {2025},
    pages     = {5344-5354}
}

@misc{egotwin25,
      title={EgoTwin: Dreaming Body and View in First Person},
      author={Jingqiao Xiu and Fangzhou Hong and Yicong Li and Mengze Li and Wentao Wang and Sirui Han and Liang Pan and Ziwei Liu},
      year={2025},
      eprint={2508.13013},
      archivePrefix={arXiv},
      primaryClass={cs.CV},
      url={https://arxiv.org/abs/2508.13013},
}

@inproceedings{ego4o25cvpr,
  title={Ego4o: Egocentric Human Motion Capture and Understanding from Multi-Modal Input},
  author={Wang, Jian and Dabral, Rishabh and Luvizon, Diogo and Cao, Zhe and Liu, Lingjie and Beeler, Thabo and Theobalt, Christian},
  booktitle=CVPR,
  pages={22668--22679},
  year={2025}
}

@misc{peva25,
        title={Whole-Body Conditioned Egocentric Video Prediction},
        author={Yutong Bai and Danny Tran and Amir Bar and Yann LeCun and Trevor Darrell and Jitendra Malik},
        year={2025},
        eprint={2506.21552},
        archivePrefix={arXiv},
        primaryClass={cs.CV},
        url={https://arxiv.org/abs/2506.21552},
}

@misc{egoppg25iccv,
      title={egoPPG: Heart Rate Estimation from Eye-Tracking Cameras in Egocentric Systems to Benefit Downstream Vision Tasks},
      author={Björn Braun and Rayan Armani and Manuel Meier and Max Moebus and Christian Holz},
      year={2025},
      eprint={2502.20879},
      archivePrefix={arXiv},
      primaryClass={cs.CV},
      url={https://arxiv.org/abs/2502.20879},
}

@inproceedings{4dgt25neurips,
    title     = {4DGT: Learning a 4D Gaussian Transformer Using Real-World Monocular Videos},
    author    = {Xu, Zhen and Li, Zhengqin and Dong, Zhao and Zhou, Xiaowei and Newcombe, Richard and Lv, Zhaoyang},
    journal   = NeurIPS,
    year      = {2025}
}

@inproceedings{wacu25iccv,
  title={WACU: Multi-Modal Wristband Assistant for Contextual Understanding},
  author={Patsch, Constantin and Goter, Jaden and Greer, Joseph and Ma, Lingni and Sodhi, Raj},
  booktitle=ICCV,
  pages={7214--7223},
  year={2025}
}

@article{chang2015shapenet,
  title={Shapenet: An information-rich 3d model repository},
  author={Chang, Angel X and Funkhouser, Thomas and Guibas, Leonidas and Hanrahan, Pat and Huang, Qixing and Li, Zimo and Savarese, Silvio and Savva, Manolis and Song, Shuran and Su, Hao and others},
  journal={arXiv preprint arXiv:1512.03012},
  year={2015}
}

@inproceedings{deitke2023objaverse,
  title={Objaverse: A universe of annotated 3d objects},
  author={Deitke, Matt and Schwenk, Dustin and Salvador, Jordi and Weihs, Luca and Michel, Oscar and VanderBilt, Eli and Schmidt, Ludwig and Ehsani, Kiana and Kembhavi, Aniruddha and Farhadi, Ali},
  booktitle={Proceedings of the IEEE/CVF conference on computer vision and pattern recognition},
  pages={13142--13153},
  year={2023}
}

@article{deitke2023objaversexl,
  title={Objaverse-xl: A universe of 10m+ 3d objects},
  author={Deitke, Matt and Liu, Ruoshi and Wallingford, Matthew and Ngo, Huong and Michel, Oscar and Kusupati, Aditya and Fan, Alan and Laforte, Christian and Voleti, Vikram and Gadre, Samir Yitzhak and others},
  journal={Advances in Neural Information Processing Systems},
  volume={36},
  pages={35799--35813},
  year={2023}
}

@article{collins2022abo,
  title={ABO: Dataset and Benchmarks for Real-World 3D Object Understanding},
  author={Collins, Jasmine and Goel, Shubham and Deng, Kenan and Luthra, Achleshwar and
          Xu, Leon and Gundogdu, Erhan and Zhang, Xi and Yago Vicente, Tomas F and
          Dideriksen, Thomas and Arora, Himanshu and Guillaumin, Matthieu and
          Malik, Jitendra},
  journal={CVPR},
  year={2022}
}

@inproceedings{downs2022google,
  title={Google scanned objects: A high-quality dataset of 3d scanned household items},
  author={Downs, Laura and Francis, Anthony and Koenig, Nate and Kinman, Brandon and Hickman, Ryan and Reymann, Krista and McHugh, Thomas B and Vanhoucke, Vincent},
  booktitle={2022 International Conference on Robotics and Automation (ICRA)},
  pages={2553--2560},
  year={2022},
  organization={Ieee}
}

@article{aanaes2016large,
  title={Large-scale data for multiple-view stereopsis},
  author={Aan{\ae}s, Henrik and Jensen, Rasmus Ramsb{\o}l and Vogiatzis, George and Tola, Engin and Dahl, Anders Bjorholm},
  journal={International Journal of Computer Vision},
  volume={120},
  number={2},
  pages={153--168},
  year={2016},
  publisher={Springer}
}

@inproceedings{liu2025uncommon,
  title={Uncommon objects in 3d},
  author={Liu, Xingchen and Tayal, Piyush and Wang, Jianyuan and Zarzar, Jesus and Monnier, Tom and Tertikas, Konstantinos and Duan, Jiali and Toisoul, Antoine and Zhang, Jason Y and Neverova, Natalia and others},
  booktitle={Proceedings of the IEEE/CVF Conference on Computer Vision and Pattern Recognition},
  pages={14102--14113},
  year={2025}
}

@inproceedings{dai2017scannet,
  title={Scannet: Richly-annotated 3d reconstructions of indoor scenes},
  author={Dai, Angela and Chang, Angel X and Savva, Manolis and Halber, Maciej and Funkhouser, Thomas and Nie{\ss}ner, Matthias},
  booktitle={Proceedings of the IEEE conference on computer vision and pattern recognition},
  pages={5828--5839},
  year={2017}
}

@inproceedings{yeshwanth2023scannet++,
  title={Scannet++: A high-fidelity dataset of 3d indoor scenes},
  author={Yeshwanth, Chandan and Liu, Yueh-Cheng and Nie{\ss}ner, Matthias and Dai, Angela},
  booktitle={Proceedings of the IEEE/CVF International Conference on Computer Vision},
  pages={12--22},
  year={2023}
}

@article{chang2017matterport3d,
  title={Matterport3d: Learning from rgb-d data in indoor environments},
  author={Chang, Angel and Dai, Angela and Funkhouser, Thomas and Halber, Maciej and Niessner, Matthias and Savva, Manolis and Song, Shuran and Zeng, Andy and Zhang, Yinda},
  journal={arXiv preprint arXiv:1709.06158},
  year={2017}
}

@article{straub2019replica,
  title={The replica dataset: A digital replica of indoor spaces},
  author={Straub, Julian and Whelan, Thomas and Ma, Lingni and Chen, Yufan and Wijmans, Erik and Green, Simon and Engel, Jakob J and Mur-Artal, Raul and Ren, Carl and Verma, Shobhit and others},
  journal={arXiv preprint arXiv:1906.05797},
  year={2019}
}

@inproceedings{roberts2021hypersim,
  title={Hypersim: A photorealistic synthetic dataset for holistic indoor scene understanding},
  author={Roberts, Mike and Ramapuram, Jason and Ranjan, Anurag and Kumar, Atulit and Bautista, Miguel Angel and Paczan, Nathan and Webb, Russ and Susskind, Joshua M},
  booktitle={Proceedings of the IEEE/CVF international conference on computer vision},
  pages={10912--10922},
  year={2021}
}

@inproceedings{fu20213d,
  title={3d-front: 3d furnished rooms with layouts and semantics},
  author={Fu, Huan and Cai, Bowen and Gao, Lin and Zhang, Ling-Xiao and Wang, Jiaming and Li, Cao and Zeng, Qixun and Sun, Chengyue and Jia, Rongfei and Zhao, Binqiang and others},
  booktitle={Proceedings of the IEEE/CVF International Conference on Computer Vision},
  pages={10933--10942},
  year={2021}
}

@article{pfaff2026scenesmith,
  title={SceneSmith: Agentic Generation of Simulation-Ready Indoor Scenes},
  author={Pfaff, Nicholas and Cohn, Thomas and Zakharov, Sergey and Cory, Rick and Tedrake, Russ},
  journal={arXiv preprint arXiv:2602.09153},
  year={2026}
}

@inproceedings{avetisyan2024scenescript,
  title={Scenescript: Reconstructing scenes with an autoregressive structured language model},
  author={Avetisyan, Armen and Xie, Christopher and Howard-Jenkins, Henry and Yang, Tsun-Yi and Aroudj, Samir and Patra, Suvam and Zhang, Fuyang and Frost, Duncan and Holland, Luke and Orme, Campbell and others},
  booktitle={European Conference on Computer Vision},
  pages={247--263},
  year={2024},
  organization={Springer}
}

@inproceedings{avetisyan2019scan2cad,
  title={Scan2cad: Learning cad model alignment in rgb-d scans},
  author={Avetisyan, Armen and Dahnert, Manuel and Dai, Angela and Savva, Manolis and Chang, Angel X and Nie{\ss}ner, Matthias},
  booktitle={Proceedings of the IEEE/CVF Conference on computer vision and pattern recognition},
  pages={2614--2623},
  year={2019}
}

@article{rao2025leveraging,
  title={Leveraging Automatic CAD Annotations for Supervised Learning in 3D Scene Understanding},
  author={Rao, Yuchen and Ainetter, Stefan and Stekovic, Sinisa and Lepetit, Vincent and Fraundorfer, Friedrich},
  journal={arXiv preprint arXiv:2504.13580},
  year={2025}
}

@inproceedings{yu2025metascenes,
  title={Metascenes: Towards automated replica creation for real-world 3d scans},
  author={Yu, Huangyue and Jia, Baoxiong and Chen, Yixin and Yang, Yandan and Li, Puhao and Su, Rongpeng and Li, Jiaxin and Li, Qing and Liang, Wei and Zhu, Song-Chun and others},
  booktitle={Proceedings of the Computer Vision and Pattern Recognition Conference},
  pages={1667--1679},
  year={2025}
}

@misc{huang2025literealitygraphicsready3dscene,
  title={LiteReality: Graphics-Ready 3D Scene Reconstruction from RGB-D Scans},
  author={Zhening Huang and Xiaoyang Wu and Fangcheng Zhong and Hengshuang Zhao and Matthias Nießner and Joan Lasenby},
  year={2025},
  eprint={2507.02861},
  url={https://arxiv.org/abs/2507.02861}
}

@misc{siddiqui2026shaper,
      title={ShapeR: Robust Conditional 3D Shape Generation from Casual Captures}, 
      author={Yawar Siddiqui and Duncan Frost and Samir Aroudj and Armen Avetisyan and Henry Howard-Jenkins and Daniel DeTone and Pierre Moulon and Qirui Wu and Zhengqin Li and Julian Straub and Richard Newcombe and Jakob Engel},
      year={2026},
      eprint={2601.11514},
      archivePrefix={arXiv},
      primaryClass={cs.CV},
      url={https://arxiv.org/abs/2601.11514}, 
}

@inproceedings{pan2023aria,
  title={Aria digital twin: A new benchmark dataset for egocentric 3d machine perception},
  author={Pan, Xiaqing and Charron, Nicholas and Yang, Yongqian and Peters, Scott and Whelan, Thomas and Kong, Chen and Parkhi, Omkar and Newcombe, Richard and Ren, Yuheng Carl},
  booktitle={Proceedings of the IEEE/CVF International Conference on Computer Vision},
  pages={20133--20143},
  year={2023}
}

@article{ravi2024sam,
  title={Sam 2: Segment anything in images and videos},
  author={Ravi, Nikhila and Gabeur, Valentin and Hu, Yuan-Ting and Hu, Ronghang and Ryali, Chaitanya and Ma, Tengyu and Khedr, Haitham and R{\"a}dle, Roman and Rolland, Chloe and Gustafson, Laura and others},
  journal={arXiv preprint arXiv:2408.00714},
  year={2024}
}

@inproceedings{zhang2024rohm,
  title={{RoHM: Robust Human Motion Reconstruction via Diffusion}},
  author={Zhang, Siwei and Bhatnagar, Bharat Lal and Xu, Yuanlu and Winkler, Alexander and Kadlecek, Petr and Tang, Siyu and Bogo, Federica},
  booktitle=CVPR,
  year={2024}
}

@misc{meta2025llama,
  title={The llama 4 herd: The beginning of a new era of natively multimodal ai innovation},
  author={Meta, AI},
  year={2025}
}
